\documentclass[preprint,3p]{elsarticle}
\usepackage{times}
\usepackage{epsfig}
\usepackage{setspace}
\usepackage{amsmath,amsthm,amssymb,amsfonts}
\usepackage{array}
\usepackage{multirow}
\usepackage{booktabs}
\usepackage{color}
\usepackage{amsfonts}
\usepackage{lineno}
\usepackage{lscape}
\usepackage{bm}
\usepackage{algpseudocode}
\usepackage{threeparttable}
\usepackage{algorithm}
\usepackage{makecell}
\usepackage{siunitx}
\usepackage{pifont}
\usepackage[colorlinks,
            linkcolor=red,
            anchorcolor=blue,
            citecolor=green
            ]{hyperref}

\biboptions{authoryear}

\journal{Journal of \LaTeX\ Templates}
\begin{document}

\begin{frontmatter}

\title{Inter- and intra-uncertainty based feature aggregation model for semi-supervised histopathology image segmentation}

\author[mymainaddress]{Qiangguo~Jin}

\author[mysecondaryaddress]{Hui~Cui}

\author[mythirdaddress]{Changming~Sun}

\author[myfourthaddress]{Yang~Song}

\author[mymainaddress]{Jiangbin~Zheng}

\author[mymainaddress]{Leilei~Cao}

\author[myfifthaddress]{Leyi~Wei}

\author[mysixthaddress]{Ran Su\corref{mycorrespondingauthor}}
\cortext[mycorrespondingauthor]{Corresponding author}

\address[mymainaddress]{School of Software, Northwestern Polytechnical University, Shaanxi, China}
\address[mysecondaryaddress]{Department of Computer Science and Information Technology, La Trobe University, Melbourne, Australia}
\address[mythirdaddress]{CSIRO Data61, Sydney, Australia}
\address[myfourthaddress]{School of Computer Science and Engineering, University of New South Wales, Sydney, Australia}
\address[myfifthaddress]{School of Software, Shandong University, Shandong, China}
\address[mysixthaddress]{School of Computer Software, College of Intelligence and Computing, Tianjin University, Tianjin, China}

\begin{abstract}
Acquiring pixel-level annotations is often limited in applications such as histology studies that require domain expertise. Various semi-supervised learning approaches have been developed to work with limited ground truth annotations, such as the popular teacher-student models. However, hierarchical prediction uncertainty within the student model (intra-uncertainty) and image prediction uncertainty (inter-uncertainty) have not been fully utilized by existing methods.
To address these issues, we first propose a novel inter- and intra-uncertainty regularization method to measure and constrain both inter- and intra-inconsistencies in the teacher-student architecture. We also propose a new two-stage network with pseudo-mask guided feature aggregation (PG-FANet) as the segmentation model. The two-stage structure complements with the uncertainty regularization strategy to avoid introducing extra modules in solving uncertainties and the aggregation mechanisms enable multi-scale and multi-stage feature integration.
Comprehensive experimental results over the MoNuSeg and CRAG datasets show that our PG-FANet outperforms other state-of-the-art methods and our semi-supervised learning framework yields competitive performance with a limited amount of labeled data.
\end{abstract}


\begin{keyword}


Semi-supervised learning \sep Feature aggregation \sep Uncertainty regularization \sep Histopathology image segmentation
\end{keyword}

\end{frontmatter}

\section{Introduction}
\label{sec:introduction}
Accurate instance segmentation in histology images is important to analyze the morphology of various structures, such as nuclei and glands, which is essential for disease diagnosis~\citep{graham2019mild}, prognostic prediction~\citep{lu2021feature}, and tissue phenotyping~\citep{javed2020cellular}. However, it is impractical to segment numerous nuclei/glands manually due to the subtle contrast between the objects of interest and background tissues, and high complexity of morphological features. Therefore, automated segmentation methods are highly demanded for histopathology images.

Deep learning based methods have demonstrated superiority in histopathology image segmentation~\citep{su2015robust,graham2019mild} and achieved outstanding performance under full supervision. 
Although these solutions have shown superior performance under full supervision, these segmentation methods rely heavily on huge numbers of pixel-level annotations, which are difficult to obtain. The difficulties in delineating histopathology images can cause extremely high workload on pathologists. Semi-supervised learning (SSL) is one of the approaches to train with limited supervision. However, semi-supervised instance segmentation remains a challenging task in histopathology image processing due to image characteristics and domain problems. First, histopathology objects, such as nuclei and glands, are often closely adjacent or overlapping with each other, making it difficult to delineate separate instances. Second, areas of uncertainty that exist within the objects and around the boundaries may not be well captured under limited supervision.

Current state-of-the-art SSL methods for biomedical image segmentation can be roughly divided in four categories. 
The first category is based on consistency regularization~\citep{tarvainen2017mean,yu2019uncertainty,wang2020double,luo2022semi,wu2022mutual,xu2023ambiguity}, which minimizes the prediction variance on a given unlabeled example and its perturbed version.
The second type is to generate pseudo labels by learning from labeled data, and then use the pseudo labels to enhance the learning from unlabeled data~\citep{li2020self,zheng2020cartilage,li2020transformation,chen2021semi,bai2023bidirectional,zhao2023rcps}.
The third one is adversarial learning, which is introduced~\citep{zhang2017deep,lei2022semi,xu2022bmanet} to enforce higher-level consistency between labeled data and unlabeled data.
The last category utilizes contrastive learning (CL)~\citep{shi2022semi,gu2022contrastive,zhang2023multi,chaitanya2023local,basak2023pseudo}, which forces the network to learn representative features in similar and dissimilar regions for segmentation.
As for histopathology image segmentation, \citet{li2020self} generated pseudo labels as guidance for nuclei segmentation. \citet{xie2020pairwise} proposed a pairwise relation-based semi-supervised ($\mathrm{PRS^2}$) model for gland segmentation on histology images. \citet{shi2022semi} and~\citet{wu2022cross} proposed contrast learning based semi-supervised model for tissue and nuclei segmentation.

However, the above models have several limitations. First, most of the previous SSL methods~\citep{li2020transformation,zhou2020deep,yu2019uncertainty} are based on the mean teacher~\citep{tarvainen2017mean} architecture. As unlabeled data do not have ground truth (GT), a common strategy is to use the predictions by the teacher model as guidance. Unfortunately, it is not guaranteed that the teacher model always generates better results than the student model from unlabeled data. Such prediction discrepancies relies crucially on the uncertainty of each target prediction by the teacher and student model (i.e., inter-uncertainty).
Second, in convolutional neural networks (CNNs), extracting features at one particular layer could affect the following layers~\citep{zheng2021rectifying,PnPAdaNet,jin2022semi}, which may cause inconsistencies during information propagation. Such inconsistencies casued by the hierarchical network are neglected. Hence, we suggest leveraging uncertainty among the student network (i.e., intra-uncertainty) to reduce such prediction inconsistencies. Third, the pseudo labels may contain noise that results in ambiguous guidances, especially along object boundaries. Finally, methods based on adversarial learning are difficult to train, and CL based methods may have issues in distinguishing the foreground pixels and background pixels in high-level feature maps due to low spatial resolution.

\begin{figure}
  \centering
  \includegraphics[scale=0.80]{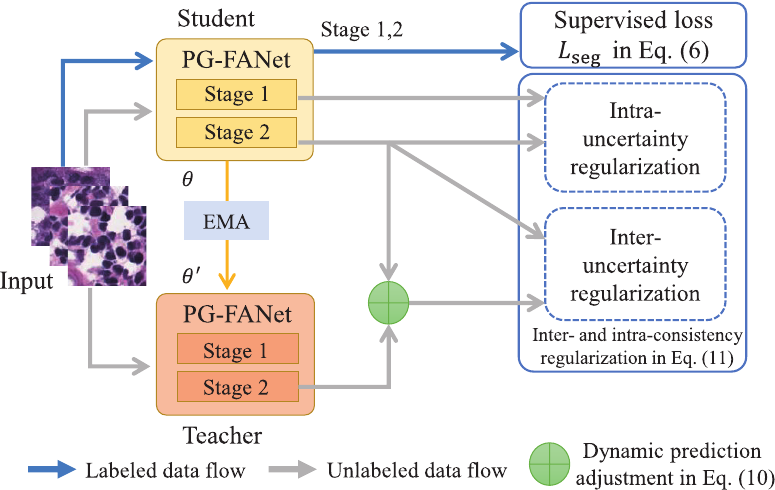}
  \caption{The overall architecture of the proposed semi-supervised histopathology image segmentation model using two-stage PG-FANet and inter- and intra-uncertainty and consistency regularization. EMA denotes exponential moving average.}
  \label{fig:overview of semi-supervised}
\end{figure}

To address these challenges existing in fully- or semi-supervised histopathology image segmentation, we propose an inter- and intra-uncertainty regularization based semi-supervised segmentation framework with a multi-scale multi-stage feature aggregation network. In our SSL setting, we gradually add fully annotated images to the labeled dataset to simulate annotating processes by pathologists.
The overall architecture is shown in Figure~\ref{fig:overview of semi-supervised}.
To learn from both labeled and unlabeled data for annotation-efficient segmentation, we exploit the learning strategy in the mean teacher framework, which is simple yet effective. In the mean teacher architecture, the student model is trained by using both labeled and unlabeled data, and the teacher model is an average of succeeding student models. The major innovations and contributions of the proposed method include:
\begin{itemize}
  \item We propose a novel pseudo-mask guided feature aggregation network (PG-FANet)\footnote{Source code will be released at https://github.com/BioMedIA-repo/PG-FANet}, which consists of two-stage sub-networks, a mask-guided feature enhancement (MGFE) module, and a multi-scale multi-stage feature aggregation (MMFA) module. The MGFE drives the attention of the network to the region of interest (ROI) under coarse semantic segmentation. The MMFA enables simultaneous aggregation of multi-scale and multi-stage features, so as to avoid the impact of feature incompatibilities in conventional U-shape skip connections.
  \item We propose an inter- and intra-uncertainty modeling and measurement mechanism to penalize both inter- and intra-uncertainties in the teacher-student architecture. The inter-uncertainty regularization formula is enhanced by a newly introduced shape attention mechanism, aiming to improve the consistency of contour-relevant predictions from the student and the teacher models.
  \item The performances of the major components in the proposed model are validated by extensive experiments over two public histopathology image datasets for nuclei and gland segmentation. Experimental results show that our PG-FANet outperforms other fully-supervised state-of-the-art models and our proposed semi-supervised learning architecture achieves desirable performance with limited labeled data when compared with recent semi-supervised learning schemes.
\end{itemize}


\section{Related Work}
\label{sec:related_works}
\subsection{Histopathology image segmentation}
Automatically segmenting histopathology images is a challenging task. Many methods have been proposed in segmenting histopathology images under full supervision. For nuclei segmentation, \citet{su2015robust}, \citet{xiang2020bio}, \citet{XIANG2022102420}, and \citet{chen2023enhancing} handled the large variations of shapes and inhomogeneous intensities of nuclei, while \citet{graham2019mild}, \citet{chen2016dcan}, and \citet{xie2019deep} proposed to reduce the ambiguity for glands of various sizes. For both tasks, a full-resolution convolutional neural network (FullNet) was proposed by \citet{qu2019improving}, where a variance constrained cross-entropy loss was introduced to explicitly learn the instance-level relation between pixels in nuclei/gland images.
\citet{yang2023ads_unet} proposed a nested U-Net that combines cascade training and AdaBoost algorithm for histopathology image segmentation.

The aforementioned methods are all fully-supervised approaches, which require pixel-level annotations by pathologists.
Considering the large number of nuclei (e.g., 28,846 in~\citep{kumar2019multi}) in a histopathology image dataset, reducing the workload of pathologists in clinical practice by leveraging reduced annotations via semi-supervised approach is worth investigation.

To deal with limited annotations, researchers attempted to exploit useful information from unlabeled data with semi-supervised techniques.
Self-loop~\citep{li2020self} was proposed to utilize the generated pseudo label as guidance to optimize the neural network.
\citet{zhou2020deep} explored a mask-guided feature distillation mechanism for nuclei instance segmentation.
\citet{xie2020pairwise} proposed a segmentation network (S-Net) and a pairwise relation network (PR-Net) for gland segmentation.
However, the vanilla pseudo labels may contain noise resulting in ambiguous guidances, and the U-Net~\citep{ronneberger2015u} based methods may introduce feature incompatibilities~\citep{ibtehaz2020multiresunet}.

In our work, we explore a pseudo label based two-stage model to help the feature representation ability for both nuclei and gland segmentation tasks.

\subsection{Feature aggregation}
Recently, deep learning networks have been proposed to enhance feature representation and aggregation in biomedical analysis tasks~\citep{yu2015multi,zhong2020squeeze,chen2018searching,yu2020context,sundaresan2021triplanar,cui2022deepmc,li2022npcnet,cao2022auto,yan2023samppred,jiang2023donet,yang2023directional,yu2023unest}. The networks for biomedical image segmentation can be roughly divided into U-shape based architectures and none-U-shape based architectures.

Since U-Net~\citep{ronneberger2015u} showed its power in dealing with biomedical image segmentation, many U-Net variations with skip-connection have been proposed for a better feature aggregation. Those methods explored the feature aggregation of the encoder and decoder to integrate global and local features~\citep{qin2020u2,ji2020uxnet,liu2019nuclei,ZHAO2020101786,qu2019improving,xu2021asymmetric,sundaresan2021triplanar,cao2022auto,yu2023unest}.
The skip connections, however, may introduce feature incompatibilities~\citep{ibtehaz2020multiresunet} and bring in discrepancies throughout the propagation.

Regarding none-U-shape based architectures, \citet{zheng2019new} proposed an ensemble learning~\citep{yan2023pretp} framework for 3D biomedical image segmentation that combined the merits of 2D and 3D models. 
\citet{zhang2019attention} proposed an attention residual learning CNN model (ARL-CNN) to leverage features at difficult stages.
\citet{li2022npcnet} introduced an automated segmentation network known as NPCNet, which comprises a position enhancement module (PEM), a scale enhancement module (SEM), and a boundary enhancement module (BEM) for the segmentation of primary nasopharyngeal carcinoma tumors and metastatic lymph nodes. \cite{jiang2023donet} presented a de-overlapping network (DoNet) within a decompose-and-recombined strategy. They aggregated rich semantic features for both overlapping and non-overlapping regions using fusion units for cytology instance segmentation.
\citet{yang2023directional} proposed a directional connectivity-based segmentation network (DconnNet) designed to separate the directional subspace from the shared latent space. They then employed the extracted directional features to enhance the overall data representation.
Considering that harnessing pseudo masks and multi-scale features to differentiate nuclei/glands from the background could benefit segmentation, we aim to propose a pseudo-mask guided feature aggregation approach. Different from previous feature aggregation work~\citep{qin2020u2,ji2020uxnet,xiang2020bio,XIANG2022102420,liu2019nuclei,ZHAO2020101786,qu2019improving}, the proposed feature aggregation incorporates pseudo-mask as guidance to attentively aggregate multi-scale features in different stages.

\subsection{Semi-supervised biomedical image segmentation} In biomedical image segmentation, different types of semi-supervised techniques have been proposed, which can be roughly categorized in four categories: (1) Consistency regularization~\citep{tarvainen2017mean,yu2019uncertainty,wang2020double,luo2022semi,wu2022mutual,xu2023ambiguity}. Learning from consistency can be regarded as learning from stability under perturbations. Numerous consistency-based methodologies have been proposed, deriving supplementary supervisory cues from unannotated data.  (2) Use of pseudo label~\citep{li2020self,zheng2020cartilage,li2020transformation,chen2021semi,bai2023bidirectional,zhao2023rcps}. Several methods have achieved robust representations through the process of acquiring supplementary information in the form of pseudo-labels. Nevertheless, owing to the inadequate class separability within the feature space, these pseudo-labels may contain potential noise and provide ambiguous guidance. (3) Adversarial learning~\citep{zhang2017deep,lei2022semi,xu2022bmanet}. The adversarial learning model aligns the distributions of segmented objects across different patients, ensuring robust predictions. However, the training process can be laborious and prone to instability. (4) Contrastive learning (CL)~\citep{shi2022semi,gu2022contrastive,zhang2023multi,chaitanya2023local,basak2023pseudo}. Recently proposed CL-based methods have demonstrated their effectiveness in distinguishing between similar and dissimilar regions within the feature space. However, it is worth noting that pixel-wise features can be challenging to differentiate, particularly in high-level feature maps with lower resolution. Among these methods, mean teacher architecture shows its learning ability and simple training procedure in consistency regularization. Hence, we introduce mean teacher based methods for a broad review.

Based on the mean teacher~\citep{tarvainen2017mean} architecture, \citet{li2020transformation} introduced a transformation consistent self-ensembling model for medical image segmentation.
\citet{yu2019uncertainty} estimated the uncertainty with the Monte Carlo dropout for semi-supervised 3D left atrium segmentation.
\citet{zhou2020deep} constrained the teacher and the student networks under mask-guided feature distillation with a perturbation-sensitive sample mining mechanism for nuclei instance segmentation.
\citet{xu2023ambiguity} selected the consistency targets to integrate informative complementary clues during training.
Nevertheless, these approaches suffer from limitations. Without the ground truth, the teacher network does not always generate better performance than the student network, which leads to misguiding segmentation.
These limitations motivate our approach. Our proposed solution leverages the uncertainties in the teacher-student architecture and constrains the inter- and intra-inconsistencies during training, so that the proposed consistency regularization strategy can be more robust.

\subsection{Uncertainty estimation}
For deep learning, Bayesian deep networks are widely used to measure prediction uncertainty~\citep{nielsen2009bayesian,kendall2017uncertainties,gal2016dropout,gustafsson2020evaluating,wang2021tripled,wang2022semi,wang2022dual,zheng2022double,zhu2023hybrid} because of its robustness and effectiveness. For example, \citet{kwon2020uncertainty} proposed to estimate the aleatoric and epistemic uncertainty in medical image classification using a Bayesian neural network.
As transformation operations in data augmentation may have an impact on segmentation results, \citet{wang2019aleatoric} analyzed the effect of such transformations by introducing a test-time augmentation-based aleatoric uncertainty.
\citet{yu2019uncertainty} generated the uncertainty map by utilizing 8 stochastic forward passes on the teacher model under random dropout.
\citet{li2020transformation} estimated uncertainties by introducing a transformation-consistent regularization strategy when ensembling models.
\citet{li2020dual} encouraged the student model to produce similar outputs as the exponential moving average (EMA) teacher model under small perturbation operations in order to eliminate the uncertainty predictions among the architecture.
\citet{wang2020double} claimed that without ground truth for unlabeled data, it cannot be guaranteed that the teacher model can provide accurate predictions. Hence, they proposed a feature-uncertainty and segmentation-uncertainty estimation method (DUW) for left atrium (LA) segmentation.
\citet{wang2021tripled,wang2022semi} introduced a foreground and background reconstruction task, along with a signed distance field (SDF) prediction task. They investigated the mutual enhancement between these two auxiliary tasks using a mean teacher architecture. Furthermore, they developed a triple-uncertainty guided framework to extract more reliable knowledge from the teacher model for medical image segmentation. 
\citet{zheng2022double} introduced a double noise mean teacher self-ensembling model for semi-supervised 2D tumor segmentation.
\citet{wang2022dual} explored multi-scale information using a dual multi-scale mean teacher network for COVID-19 segmentation.
\citet{zhu2023hybrid} introduced a hybrid dual mean teacher (HD-Teacher) model that incorporates hybrid, semi-supervised, and multi-task learning techniques to achieve semi-supervised segmentation of MRI scans.
In addition to the prediction variances from the teacher and student models, the intra-uncertainties caused by the hierarchical CNN architecture are neglected by previous work, which motivated our approach. Inspired by the works mentioned above, we leverage the intra-uncertainties in the teacher-student network by enforcing the small perturbations at extra feature-level in our novel segmentation model.

\section{Methodology}
\label{sec:methodology}

\subsection{Problem formulation}
Given a labeled dataset $(\mathcal{X}_{\mathrm{l}},\mathcal{Y}_{\mathrm{l}}) =\left\{(x_{i},y_{i})\right\}_{i=1}^{M}$ and an unlabeled dataset $\mathcal{X}_{\mathrm{u}} =\left\{(x_{i})\right\}_{i=1}^{N}$, where $M$ is the number of images with known segmentation results, each image $x_{i}$ has a corresponding segmented mask $y_{i}$, $N$ denotes the number of unlabeled images, and $\mathcal{X}_{\mathrm{u}}$ represents images without labeled masks during the training process.
The segmentation task aims to learn a mapping function $F$ from input images $\mathcal{X}$ to segmentation $\mathcal{Y}$. In semi-supervised learning, the parameters $\theta$ of $F$ are optimized on the labeled and the unlabeled datasets as follows:
\begin{equation}
  \begin{split}
    \min _{\theta} \sum_{i=1}^{M} L_{\mathrm{seg}}\left(F\left(x_{i} | \theta\right), y_{i}\right)+\lambda L_{\mathrm{c}}(\theta, (\mathcal{X}_{\mathrm{l}},\mathcal{Y}_{\mathrm{l}}), \mathcal{X}_{\mathrm{u}}),
  \end{split}
  \label{eq:semi_supervised_learning_formulation}
\end{equation}
where $L_{\mathrm{seg}}$ is the supervised loss function, $L_{\mathrm{c}}$ is the unsupervised consistency loss, and $\lambda$ is a weighting factor to enforce the consistency between the two datasets.
As discussed before, the intra-uncertainty of the student model is neglected in the recent teacher-student network during the learning process.  Furthermore, without ground truth labels for unlabeled data, it is difficult to determine whether the teacher model provides more accurate results than the student model or not.

To address these issues, we propose an uncertainty modeling mechanism to measure the intra-model uncertainties within the student network and the inter-model uncertainties between the student and teacher networks. Accordingly, we propose an inter-uncertainty consistency loss ($L_{\mathrm{inter}}$) and a new intra-uncertainty penalization term ($L_{\mathrm{intra}}$), and the learning process and overall loss function in Equation (\ref{eq:semi_supervised_learning_formulation}) are revised as:
\begin{equation}
  \begin{split}
    &\min _{\theta} \!\sum_{i=1}^{M} L_{\mathrm{seg}}\!\left(F\left(x_{i} | \theta\right) , y_{i}\right) \\
    & + \lambda (t) (L_{\mathrm{inter}}(\theta, (\mathcal{X}_{\mathrm{l}},\mathcal{Y}_{\mathrm{l}}), \mathcal{X}_{\mathrm{u}})  + \lambda_{\mathrm{intra}} L_{\mathrm{intra}}),
  \end{split}
  \label{eq:lr_semi_supervised_learning_formulation}
\end{equation}
where $L_{\mathrm{inter}}$ denotes the unsupervised consistency loss for minimizing the inter-uncertainty, $L_{\mathrm{intra}}$ represents the additional regularization, which incorporates intra-uncertainty into the optimization objective to model uncertainties, $\lambda (t)$ denotes the step-related ramp-up weight factor for the consistency loss, $t$ represents the $t$th training step, and $\lambda_{\mathrm{intra}}$ is a weight factor to control the regularization.

\begin{figure*}
    \centering
    \includegraphics[scale=0.56]{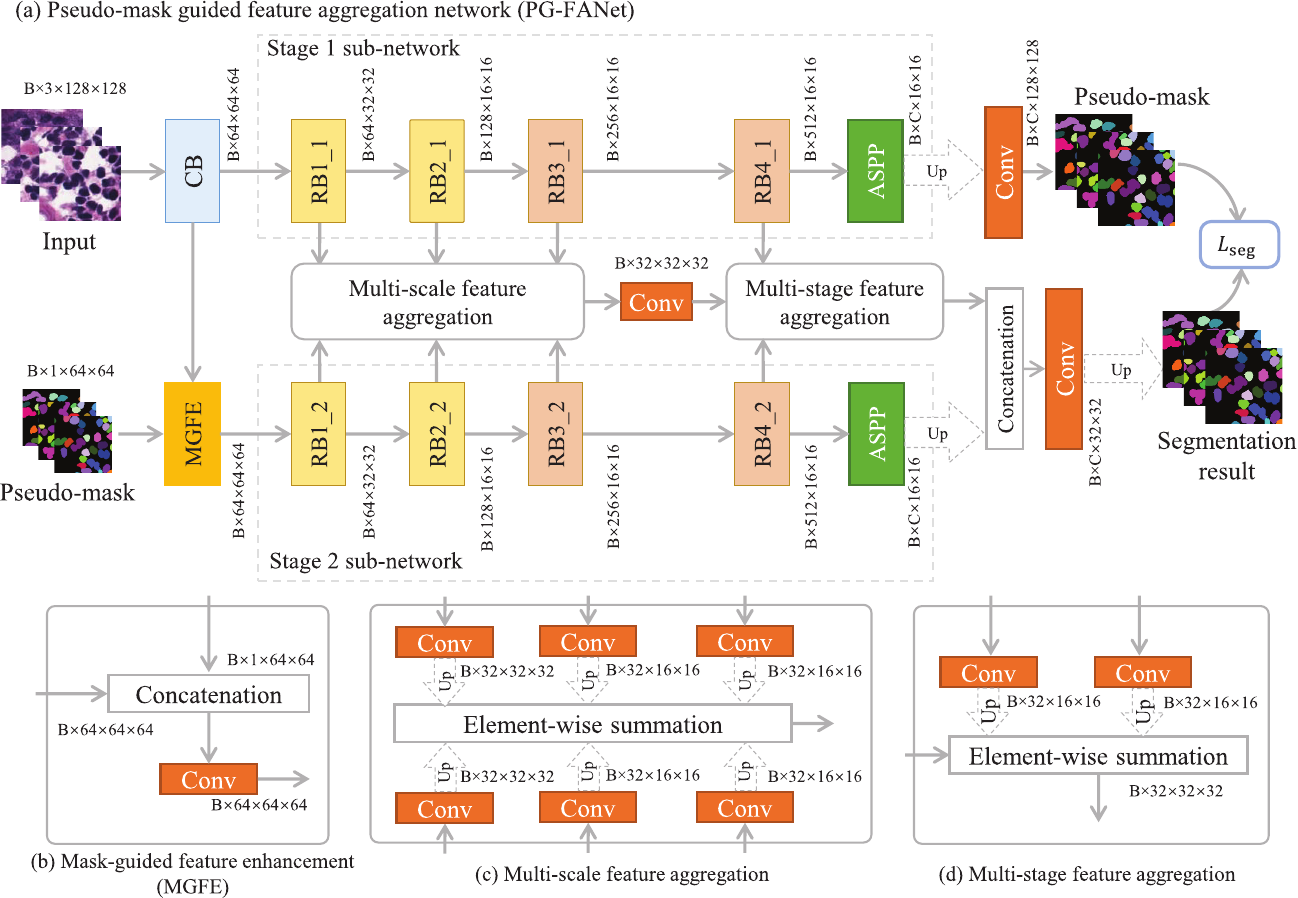}
    \caption{Overview of our (a) PG-FANet with two-stage sub-networks, (b) mask-guided feature enhancement (MGFE) module, (c) multi-scale feature aggregation, and (d) multi-stage feature aggregation. Both stages share the same convolution block (CB).
    RB$i\_s$ denotes the $i$th residual blocks in stage $s$. RB3$\_s$ and RB4$\_s$ are dilated residual blocks with dilation rates of 2 and 4 respectively to generate feature maps with various receptive fields. Conv denotes the convolutional layer. ASPP denotes the atrous spatial pyramid pooling module. Up denotes the upsampling operation. The size of the output feature maps is given by batch size $\times$ channel size $\times$ height $\times$ width (B $\times$ C $\times$ H $\times$ W).
    }
    \label{fig:overview of architecture}
  \end{figure*}
\subsection{Pseudo-mask guided feature aggregation network (PG-FANet)}
\label{sec:PG-FANet}
For effectively improving learning ability, we propose a feature aggregation network for both supervised and semi-supervised learning processes. The architecture of PG-FANet is illustrated in Figure~\ref{fig:overview of architecture}. The PG-FANet consists of three major components, i.e., two-stage sub-networks, mask-guided feature enhancement (MGFE) modules, and multi-scale multi-stage feature aggregation (MMFA) modules. MGFE is to force the attention of the network to the ROI under the guidance of the semantic pseudo-mask at feature level. MMFA is designed to extract and aggregate multi-scale and multi-stage features simultaneously to avoid the problems of feature incompatibilities in the U-shape skip connections~\citep{ibtehaz2020multiresunet}.
The final output of our PG-FANet is obtained by fusing the output of the second stage and the aggregated features.

\subsubsection{Two-stage sub-networks and MGFE}
As shown in Figure~\ref{fig:overview of architecture}, the first stage is for coarse pseudo-mask generation and the second stage is for refinement. The two-stage sub-networks follow the same architecture where each stage consists of four residual blocks (RB), an atrous spatial pyramid pooling (ASPP)~\citep{chen2017deeplab} module, and a final convolutional layer. The output feature maps of the convolution block (CB) progressively flow to the second stage. Those features and the pseudo mask of a sub-network in an early stage are then concatenated as an input to a later stage sub-network. Afterwards, the mask-guided features are adapted by a $1 \times 1$ convolutional layer for further propagation as shown in Figure~\ref{fig:overview of architecture}(b). In this way, the pseudo masks can serve as feature selectors to extract features from object instances with various sizes and shapes.

\subsubsection{MMFA}
Given the extracted features at different scales, shapes, and densities from the two-stage sub-networks, we use the MMFA module for the aggregation of multi-scale and multi-stage features. 


\textit{Multi-scale feature aggregation.}
Multi-scale feature aggregation (Figure~\ref{fig:overview of architecture}(c)) is to combine low-level features of each stage. The aggregation is formulated as Equation~(\ref{eq:multi_scale_aggregation}). For the $s$th stage, the $i$th RB can be defined as $\phi^{i}_{s}(\cdot)$, where $s$ is in $\{1, ..., S\}$ and $i$ is in $\{1, ..., I\}$. The multi-scale feature aggregation process can be formulated as:
\begin{equation}
  \begin{split}
    \mathbf{X_{m}}=\sum_{s=1}^{S} \sum_{i=1}^{I} \mathrm{Up}\left(\delta\left(\mathcal{B}(\mathrm{Conv}(\phi^{i}_{s}(\mathbf{X}^{i-1}_{s})))\right)\right),
  \end{split}
  \label{eq:multi_scale_aggregation}
\end{equation}
where $\mathbf{X_{m}}$ represents the aggregated feature map, $\mathrm{Up}$ is the upsampling operation, $\delta$ denotes the parametric rectified linear unit (PReLU)~\citep{he2015delving}, $\mathcal{B}$ denotes batch normalization (BN)~\citep{ioffe2015batch}, $\mathrm{Conv}$ represents the convolutional layer, and $\mathbf{X}^{i-1}_{s}$ denotes the output feature maps from the ($i-1$)th RB in the $s$th stage. In this work, $S$ is set to 2, and $I$ is set to 3. The multi-scale feature aggregation process reuses the mask-guided information and gains a better representation for further propagation.

\textit{Multi-stage feature aggregation.}
As the network goes deeper, low-level features and spatial information such as region boundaries may be lost~\citep{li2019dfanet}. Instead of introducing U-shape skip connections, which suffer from feature incompatibilities~\citep{ibtehaz2020multiresunet}, we fuse the early-stage features and the later-stage semantic features using the multi-stage feature aggregation module (Figure~\ref{fig:overview of architecture}(d)). The early-stage features before the ASPP and convolutional layers are more label-like ones, and the aggregation of those features would contribute to a more accurate and robust result.
Formally, the multi-stage feature aggregation can be computed as:
\begin{equation}
  \begin{split}
    \mathbf{X_{h}}=\mathbf{X_{m}^{'}} + \sum_{s=1}^{S} \mathrm{Up}\left(\delta\left(\mathcal{B}(\mathrm{Conv}(\phi^{4}_{s}(\mathbf{X}^{3}_{s})))\right)\right),
  \end{split}
  \label{eq:high_level_aggregation}
\end{equation}
where $\mathbf{X_{m}^{'}}$ denotes the fused features after the multi-scale aggregation operation, and $\mathbf{X_{h}}$ represents the output feature map after multi-stage aggregation.

\subsubsection{Loss function for PG-FANet}
As our PG-FANet has two-stage sub-networks, we optimize each stage by the standard cross-entropy loss $L_{\mathrm{ce}}$ and the Dice loss  $L_{\mathrm{Dice}}$. Hence, the main loss of each stage for supervised training is $L_{\mathrm{ce}}+L_{\mathrm{Dice}}$. The cross-entropy loss, Dice loss, or the combination of them are commonly used in many medical image segmentation methods and achieve remarkable success~\citep{wong20183d}. However, histopathology image segmentation requires not only to segment the nuclei/glands from the background, but also to separate each individual object from others. To consider the spatial relationship between objects, we also exploit the variance constrained cross ($\mathrm{vcc}$)~\citep{qu2019improving} loss as an auxiliary loss in the final stage for refined segmentation. Formally, given a minibatch $B \subseteq \mathcal{X}_{\mathrm{l}}$, the $L_{\mathrm{vcc}}$ loss is defined as~\citep{qu2019improving}:
\begin{equation}
  \begin{split}
    L_{\mathrm{vcc}}=\frac{1}{D} \sum_{d=1}^{D} \frac{1}{\left|B_{d}\right|} \sum_{i=1}^{\left|B_{d}\right|}\left( \mu_{d}-p^{i}\right)^{2}
  \end{split}
  \label{eq:varCE}
\end{equation}
where $D$ is the number of instances in $B$, $B_{d}$ denotes all the pixels that belong to instance $d$ in the minibatch, $\left|B_{d}\right|$ represents the number of pixels in $B_{d}$, $\mu_{d}$ is the mean value of the probabilities of all the pixels in $B_{d}$, and $p^{i}$ is the probability of the correct class for a pixel $i$ in the final stage.
In summary, $L_{\mathrm{seg}}$ in Equation~(\ref{eq:lr_semi_supervised_learning_formulation}) for supervised segmentation is defined as:
\begin{equation}
  \begin{split}
    L_{\mathrm{seg}} = L_{\mathrm{ce}}+\lambda_{\mathrm{Dice}}L_{\mathrm{Dice}}+\lambda_{\mathrm{vcc}} L_{\mathrm{vcc}},
  \end{split}
  \label{eq:semgnetation_loss}
\end{equation}
where $\lambda_{\mathrm{Dice}}$ and $\lambda_{\mathrm{vcc}}$ are parameters for adjusting weights.

\subsection{Inter- and intra-uncertainty and consistency regularization}
\label{sec:semi_supervised_network}
In a teacher-student framework, the teacher model is to evaluate images under perturbations for better performances~\citep{li2020dual,li2020transformation}. The weights $\theta_{t}^{\prime}$ of the teacher model at training step $t$ are updated as $\theta_{t}^{\prime}=\alpha \theta_{t-1}^{\prime}+(1-\alpha) \theta_{t}$ by the exponential moving average (EMA) weights of the student model $F$ with weights $\theta_{t}$. $\alpha$ is the EMA decay rate to update the weights of the student model with gradient descent in a total of $T$ training steps. Although these methods model the uncertainty between the student and the teacher models to some extent, the uncertainty within the student network is neglected. Due to the hierarchical architecture of the convolutional networks, the features heavily rely on those from previous layers. Thus, discrepancies may exist within the student model. Figure~\ref{fig:uncertainty_calcu} shows prediction inconsistencies that exist near region boundaries, and the network contains ambiguous predictions between two stages. Additionally, without ground truth for $\mathcal{X}_{\mathrm{u}}$, accurate prediction by the teacher model cannot always be guaranteed~\citep{wang2020double}. To address this issue, we propose a novel uncertainty and consistency regularization strategy to quantify inter- and intra-uncertainties and inconsistencies in the teacher-student architecture. We hypothesize that the inter- and intra-consistencies can provide stronger regularization, thereby enabling a better performed student network.

\subsubsection{Inter- and intra-uncertainty estimation}
To address the inconsistency issue, previous works~\citep{li2020transformation,li2020dual} model the uncertainty via inter-prediction variance of the student and the teacher networks as:
\begin{equation}
  \begin{split}
    U_{\mathrm{inter}}=\sum_{i=1}^{N} \mathbb{E}\left\|F_{s2}\left(x_{i} | \theta\right)-F_{s2}\left(x_{i} | \theta^{\prime}\right)\right\|^{2},
  \end{split}
  \label{eq:external_prediction}
\end{equation}
where $F_{s2}\left(x_{i} | \theta\right)$ represents the output of the second stage of the student model, and $F_{s2}\left(x_{i} | \theta^{\prime}\right)$ denotes the output of the second stage of the teacher model in our work.
However, intra-discrepancy exists in the different stages within the student network because of the hierarchical architecture, which will result in inconsistent predictions from the output of different stages. To remedy discrepancies, the resulting predictions from the output of each stage must be extremely consistent. Thus, we additionally estimate the intra-uncertainty ($U_{\mathrm{intra}}$) as:
\begin{equation}
  \begin{split}
    U_{\mathrm{intra}}= \sum_{i=1}^{N} \mathbb{E}\left\| F_{s1}\left(x_{i} | \theta\right) - F_{s2}\left(x_{i} | \theta\right)\right\|^{2},
  \end{split}
  \label{eq:external_prediction_ie}
\end{equation}
where $F_{s1}\left(x_{i} | \theta\right)$ denotes the output of the first stage of the student model as shown in Figure~\ref{fig:overview of architecture}(a). To enforce consistency and improve the robustness, we further introduce small perturbations to hidden features in the two-stage of the student model.

On the one hand, the supervised learning process continuously improves the ability of the two-stage student model according to Equation~(\ref{eq:semgnetation_loss}). On the other hand, the semi-supervised learning process forces the final prediction of the student model to be consistent with that of the teacher model (Equation~(\ref{eq:external_prediction})). At the same time, the student model pushes the prediction of the first stage to be consistent with its final prediction (Equation~(\ref{eq:external_prediction_ie})). With such operations, the prediction of the first stage could be more accurate, which benefits the final prediction.

\subsubsection{Inter- and intra-consistency regularization}
As aforementioned, the teacher model cannot be guaranteed to produce more accurate predictions than the student model. To dynamically prevent the teacher model from obtaining high uncertainty estimation, inspired by~\citet{wang2020double}, we introduce a learnable loss function for penalizing the uncertainty generated by the teacher model. Firstly, we calculate the uncertainty of the $i$th image in the unlabeled dataset as:
\begin{equation}
  \begin{split}
    u_{i}^{\prime}=- \hat{q}_{\mathrm{tea}}^{i} \log \hat{q}_{\mathrm{tea}}^{i},
  \end{split}
  \label{eq:uncertainty}
\end{equation}
where $\hat{q}_{\mathrm{tea}}^{i}$ is the probability prediction provided by the teacher model, and $u_{i}^{\prime}$ represents the rectified uncertainty of the prediction on the $i$th sample. Secondly, we dynamically adjust the prediction of the teacher model as follows:
\begin{equation}
  \begin{split}
    \hat{q}_{\mathrm{tea}}^{i\prime}=\left(1- u_{i}^{\prime}\right)\hat{q}_{\mathrm{tea}}^{i}+u_{i}^{\prime} \hat{q}_{\mathrm{stu}}^{i},
  \end{split}
  \label{eq:uncertainty_adjust}
\end{equation}
where $\hat{q}_{\mathrm{tea}}^{i}$ and $\hat{q}_{\mathrm{stu}}^{i}$ denote the final predictions of the teacher model and the student model respectively, $\hat{q}_{\mathrm{tea}}^{i\prime}$ represents the learnable prediction logits of the teacher model. When the teacher model provides unreliable results (high uncertainty), $\hat{q}_{\mathrm{tea}}^{i\prime}$ is approximating $\hat{q}_{\mathrm{stu}}^{i}$. On the contrary, when the teacher model is confident (with low uncertainty), $\hat{q}_{\mathrm{tea}}^{i\prime}$ remains the same with $\hat{q}_{\mathrm{tea}}^{i}$, and certain predictions are provided as targets for the student model to learn from.
Finally, we incorporate the inter- and intra-uncertainty (i.e., $U_{\mathrm{inter}}$ and $U_{\mathrm{intra}}$) into the training objective. The loss function of inter- and intra-consistency regularization can be formulated as:
\begin{equation}
  \begin{split}
    L_{\mathrm{inter}} = & L_{\mathrm{mse}} \left(F_{s2}(x_{i} | \theta), F_{s2}^{\prime}(x_{i} | \theta^{\prime})\right),\\
    L_{\mathrm{intra}} =  & L_{\mathrm{mse}} (F_{s1}(x_{i} | \theta), F_{s2}(x_{i} | \theta)),
  \end{split}
  \label{eq:uncertainty_adjust_loss_func}
\end{equation}
where $F_{s2}^{\prime}(x_{i} | \theta^{\prime})$ denotes the learnable prediction of the teacher model in Equation~(\ref{eq:uncertainty_adjust}), $L_{\mathrm{mse}}$ denotes the mean square error loss function, and $L_{\mathrm{inter}}$ and $L_{\mathrm{intra}}$ are the terms in Equation~(\ref{eq:lr_semi_supervised_learning_formulation}).
\subsubsection{Shape attention weighted consistency regularization}
In addition to promoting the complete segmentation for histopathology images, we leverage shape information to enhance the attention to boundary regions for better segmentation prediction. The shape consistency enhancement is crucial, given the observation that the increase of uncertainty in medical images mainly comes from ambiguous boundary regions. In this regard, we propose an attentive shape weight ($U_{\mathrm{shape}}$) to explicitly promote the contour-relevant predictions from the student and the teacher models. The $U_{\mathrm{shape}}$ can be formulated as:
\begin{equation}
  \begin{split}
    u_{\mathrm{shape}} & =\left\|\mathrm{Softmax}(F_{s2}(x_{i} | \theta))-\mathrm{Softmax}(F_{s2}(x_{i} | \theta^{\prime}))\right\|, \\
     U_{\mathrm{shape}} & =  - u_{\mathrm{shape}} \log u_{\mathrm{shape}}.
  \end{split}
  \label{eq:boundary_attention}
\end{equation}
As the final stage produces more accurate predictions than the first stage, additional shape information could benefit final predictions. Thus, we fuse the shape attention to $L_{\mathrm{inter}}$. Finally, the inter-consistency regularization with shape attention weights can be defined as:
\begin{equation}
  \begin{split}
    L_{\mathrm{inter}} = (1+\sigma(U_{\mathrm{shape}})) L_{\mathrm{mse}} \left(F_{s2}(x_{i} | \theta), F_{s2}^{\prime}(x_{i} | \theta^{\prime})\right),\\
  \end{split}
  \label{eq:attentive_boundary_consistency}
\end{equation}
where $\sigma$ denotes the min-max normalization to scale the uncertainty to [0,1]. In this way, differences in boundary predictions between student and teacher models can be weighted by a shape attention mechanism, allowing more targeted adjustments during training to preserve complete shapes of segmented objects.

\section{Experiments}
\label{sec:exp_and_res}
\subsection{Datasets and pre-processing}
\textbf{MoNuSeg:} The multi-organ nuclei segmentation dataset~\citep{kumar2019multi} consists of 44 H\&E stained histopathology images, which are collected from multiple hospitals.
These images are of $1000 \times 1000$ pixel resolution. The training set contains 30 histopathological images with
hand-annotated nuclei, while the test set consists of 14 images. We use the randomly sampled 27 images as training data. The remaining 3 images are used as validation data.

\textbf{CRAG:} There are 213 H\&E CRA images with different cancer grades taken from 38 whole slide images (WSIs) in the colorectal adenocarcinoma gland (CRAG) dataset~\citep{awan2017glandular}. The CRAG dataset is split into 173 training images, where 153 images are used for training and 20 images are used for validation, and 40 test images. Most of the images are of $1512 \times 1516$ pixel resolution with instance-level ground truth.

We crop patches from each training image using a sliding window. For MoNuSeg, the patch size is $128 \times 128$, resulting in 1,728 patches.
For CRAG, we extract 5,508 patches with $480 \times 480$ pixels from the 153 images.
We further perform online data augmentations including random scale, flip, rotation, and affine operations.
All these training images are normalized by using the mean and standard deviation for the images in ImageNet~\citep{deng2009imagenet}.

\subsection{Experimental settings and parameters}

We implement PG-FANet in PyTorch. Our experiments are conducted using an NVIDIA GeForce RTX 3090 graphics card. The batch size is set as 16 for MoNuSeg and 4 for CRAG. The Adam optimizer is applied with a learning rate at $\num{2.5e-4}$. The optimizer is used with a polynomial learning rate policy, where the initial learning rate is multiplied by $\left(1-\frac{iter}{total\_{iter}}\right)^{power}$ with $power$ fixed at 0.9. The total number of training iterations is set to $300* (iter\_per\_epoch)$, which is equivalent to 300 epochs as introduced in~\citet{jin2022semi}.
$\lambda_{\mathrm{Dice}}$ and $\lambda_{\mathrm{vcc}}$ in Equation~(\ref{eq:semgnetation_loss}) are both set to 1. $\lambda_{\mathrm{intra}}$ is empirically set to 1 in Equation~(\ref{eq:lr_semi_supervised_learning_formulation}) . The weight factor $\lambda (t)$ in Equation~(\ref{eq:lr_semi_supervised_learning_formulation}) is calculated by a Gaussian ramp-up function $\lambda(t)=k * e^{\left(-5(1-t/T)^{2}\right)}$, as introduced in~\citet{li2020transformation}. $T$ is set to be equivalent to the total training epoch 300, and $k$ is set to 0.1/5.0 for the MoNuSeg/CRAG dataset empirically. The EMA decay rate $\alpha$ is set to 0.99 empirically. We use the ImageNet pre-trained ResNet34~\citep{he2016deep} as backbone. All the performances are averages over 3 runs. For semi-supervised learning, we use the same settings as above, except for the growing percentage of training samples. The percentages of training images are 5\%~(1/8), 10\%~(3/15), 20\%~(5/31), and 50\%~(14/76) for nuclei/gland.

\begin{table}[]
  \centering
  \begin{threeparttable}
    \caption{Performance comparison of the proposed PG-FANet and state-of-the-art methods on the MoNuSeg dataset.}
    \label{table:monuseg_performance_comp_using_full_data}
    \renewcommand\arraystretch{0.8}
    \renewcommand\tabcolsep{9pt}
    \begin{tabular}{c|ccccc|c}
      \toprule
      \textbf{Method}                                         & \textbf{F1}    & \textbf{Dice}    & \textbf{IoU} & \textbf{AJI} & \textbf{95HD} &\textbf{Params(M)} \\ \hline
      Micro-Net~\citep{raza2019micro}      & 0.810   &0.723 & 0.602   & 0.457  & 9.753 &186.74 \\
      U2-Net~\citep{qin2020u2}             & 0.886  &0.813 & 0.704    &0.598   & 6.747  & \textbf{1.14}\\
      R2U-Net~\citep{alom2018nuclei}       & 0.866  & 0.824  & 0.718  &0.593  & 6.127   &39.09 \\
      LinkNet~\citep{chaurasia2017linknet} & 0.892 &0.825  &0.718 & 0.614  & 6.093 & 11.53\\
      FullNet~\citep{qu2019improving} &  0.897  &0.827 & 0.722  &  0.625  &5.876 &1.78 \\
      BiO-Net~\citep{xiang2020bio}    & 0.894 & 0.824 & 0.720  &0.619  & 6.008 &14.97\\
      MedFormer~\citep{gao2022data}    & 0.891 & 0.829 & 0.725  &0.629  & 5.951  &28.07\\
      HARU-Net~\citep{chen2023enhancing}    & 0.895 & 0.829 & 0.723  &0.624  & 5.964 &44.08\\
      ADS$\_$UNet~\citep{yang2023ads_unet}    & 0.894 & 0.831 & 0.728  &0.619  & 5.885& 26.72 \\\hline
      Micro-Net\tnote{*}~~\citep{raza2019micro} & -  &0.819  & 0.696   & -  & - &186.74\\
      M-Net\tnote{*}~~\citep{mehta2017m} & - &0.813  & 0.686    & -  & - &1.56\\
      R2U-Net\tnote{*}~~\citep{alom2018nuclei}  & - & 0.815 & 0.683    &-  &-&9.09    \\
      LinkNet\tnote{*}~~\citep{chaurasia2017linknet} &- &0.767  &0.625 &-  & - &11.53 \\
      FullNet\tnote{*}~~\citep{qu2019improving} & 0.857 &0.802 &-   & 0.600  & -&1.78  \\
      BiO-Net\tnote{*}~~\citep{xiang2020bio}    & - & 0.825  & 0.704  &-  & - &14.97 \\
      BiX-NAS\tnote{*}~~\citep{XIANG2022102420} & -  & 0.822& 0.699   &-  & -&-\\  \hline
      \textbf{PG-FANet}  &  \textbf{0.900}  & \textbf{0.839}&\textbf{0.736}   &\textbf{ 0.645} &\textbf{5.420}  &42.78\\
      \bottomrule
    \end{tabular}
    \begin{tablenotes}
      \footnotesize
      \item[*] Copied directly from original papers
    \end{tablenotes}
  \end{threeparttable}
\end{table}

\subsection{Comparison methods and evaluation metrics}

\subsubsection{Comparison methods}
To demonstrate the effectiveness of our proposed method, we compare our approach with several fully- or semi-supervised segmentation models. We choose the following representative methods in natural scene segmentation, nuclei/gland segmentation, and other medical image segmentation applications. 

\textbf{Fully supervised methods}: The fully-supervised methods can be categorized as U-Net based models and none-U-Net based models: (1) For U-Net based models, we choose R2U-Net~\citep{alom2018nuclei}, BiO-Net~\citep{xiang2020bio}, BiX-NAS~\citep{XIANG2022102420}, M-Net~\citep{mehta2017m}, HARU-Net~\citep{chen2023enhancing}, and ADS$\_$UNet~\citep{yang2023ads_unet} for comparison. (2) For none-U-Net based models, LinkNet~\citep{chaurasia2017linknet}, Micro-Net~\citep{raza2019micro}, FullNet~\citep{qu2019improving}, DCAN~\citep{chen2016dcan}, MILD-Net~\citep{graham2019mild}, DSE~\citep{xie2019deep}, $\mathrm{PRS^2}$~\citep{xie2020pairwise}, and MedFormer~\citep{gao2022data} are selected.

\textbf{Semi-supervised methods}: The number of published peer-reviewed SSL methods for nuclei/gland segmentation is relatively small, and we choose Self-loop~\citep{wang2020double} and $\mathrm{PRS^2}$~\citep{xie2020pairwise} for comparison. Apart from that, we find that the recently proposed consistency regularization based semi-supervised learning models in other medical image segmentation approaches can be adapted (e.g., UA-MT~\citep{yu2019uncertainty}, ICT~\citep{verma2019interpolation}, TCSM~\citep{li2020transformation}, DUW~\citep{wang2020double}, CPS~\citep{chen2021semi}, URPC~\citep{luo2022semi}, and MC-Net~\citep{wu2022mutual}).

\subsubsection{Evaluation metrics}
Evaluation measures for nuclei segmentation include F1-score (F1), intersection over union (IoU), average Dice coefficient (Dice), aggregated Jaccard index (AJI), and 95\% Hausdorff distance (95HD)  as introduced in~\citep{xiang2020bio,kumar2019multi}, and~\citet{liu2019nuclei}.
For gland segmentation, F1, object-level Dice coefficient ($\mathrm{Dice_{obj}}$), object-level Hausdorff distance ($\mathrm{Haus_{obj}}$), and 95\% object-level Hausdorff distance ($\mathrm{95HD_{obj}}$) are used for detailed evaluation as described in~\citet{chen2016dcan,graham2019mild}, and~\citet{xie2019deep}.

\subsection{Comparison with fully supervised state-of-the-arts}
Following existing literature, we firstly conduct experiments on PG-FANet with full supervision, termed as PG-FANet, on the MoNuSeg and CRAG datasets.
On the one hand, we re-implement some of existing models on the MoNuSeg and CRAG datasets under the same experimental settings for fair comparisons. On the other hand, for those complex models without source codes, we directly copy the values from their original papers. All the re-implemented models follow the same data augmentation and training strategies.

\begin{table}[]
  \centering
  \begin{threeparttable}
    \caption{Performance comparison of the proposed PG-FANet and state-of-the-art methods on the CRAG dataset.}
    \label{table:crag_performance_comp_using_full_data}
    \renewcommand\arraystretch{0.9}
    \renewcommand\tabcolsep{12.0pt}
    \begin{tabular}{c|cccc|c}
      \toprule
      \textbf{Method}  & \textbf{F1}    & \textbf{$\mathrm{\textbf{Dice}_{obj}}$} & \textbf{$\mathrm{\textbf{Haus}_{obj}}$} &  \textbf{$\mathrm{\textbf{95HD}_{obj}}$}&\textbf{Params (M)} \\ \hline
      U-Net~\citep{ronneberger2015u}                & 0.733&0.832&188.031 &  160.567&3.35 \\
      MedFormer~\citep{gao2022data}    & 0.813 & 0.885 & 118.204  &96.861  &28.07\\
      HARU-Net~\citep{chen2023enhancing}    & 0.841 & 0.875 & 137.774  &107.728 &44.08 \\
      ADS$\_$UNet~\citep{yang2023ads_unet}    & 0.749 & 0.835 & 164.886  &143.721 & 26.72 \\\hline
      DCAN\tnote{*}~~\citep{chen2016dcan}                & 0.736          & 0.794                                   & 218.760     & -& \textbf{1.75} \\
      MILD-Net\tnote{*}~~\citep{graham2019mild}          & 0.825          & 0.875                                   & 160.140    &  - &  55.69 \\
      DSE\tnote{*}~~\citep{xie2019deep}                  & 0.835          & 0.889                                   & 120.127   &  -  & -\\
      $\mathrm{PRS^2}$\tnote{*}~~\citep{xie2020pairwise} & 0.843          & 0.892                                   & 113.100    &  - & -  \\ \hline
      \textbf{PG-FANet}                  & \textbf{0.860}&	\textbf{0.901}&	\textbf{102.683}&	\textbf{80.181}&42.78 \\
      \bottomrule
    \end{tabular}
    \begin{tablenotes}
      \footnotesize
      \item[*] Copied directly from original papers
    \end{tablenotes}
  \end{threeparttable}
\end{table}

\begin{table*}[]
  \centering
  \caption{Experimental results on MoNuSeg and CRAG using our PG-FANet and state-of-the-art semi-supervised learning methods with different percentages of labeled data. Note that 5\% (1/8) denotes 5\% labeled data and the corresponding number of labeled data is 1/8 for MoNuSeg/CRAG dataset.}
  \label{table:performance_comp_using_semi_data}
  \renewcommand\arraystretch{0.8}
  \renewcommand\tabcolsep{2.0pt}
  \begin{tabular}{cc|ccccc|cccc}
    \toprule
    \multirow{2}{*}{\textbf{Labeled data}}
    & \multirow{2}{*}{\textbf{Methods} } & \multicolumn{5}{c|}{\textbf{MoNuSeg}} & \multicolumn{4}{c}{\textbf{CRAG}} \\ \cline{3-11}
                             &                                    & \textbf{F1}                           & \textbf{Dice}                      & \textbf{IoU}  & \textbf{AJI}  & \textbf{95HD} & \textbf{F1}    & \textbf{$\mathrm{\textbf{Dice}_{obj}}$} & \textbf{ $\mathrm{\textbf{Haus}_{obj}}$} & \textbf{$\mathrm{\textbf{95HD}_{obj}}$}  \\ \hline
    \multirow{6}{*}{ 5\% (1/8)  }  
    & PG-FANet  & 0.822& 0.767&	0.646&0.505&8.998& 0.718&	0.773&	246.130&	208.665  \\
    & MT~\citep{tarvainen2017mean}& 0.785&	0.790& 	0.677&0.486&	8.125 & 0.735&	0.827&	168.896&145.527 \\
    & UA-MT~\citep{yu2019uncertainty} &0.804&	0.791&	0.678& 0.498&	8.090 &0.719&0.804&194.210&162.306  \\
    & ICT~\citep{verma2019interpolation}  &0.789&0.793&	0.680&	0.495&	7.846  & 0.646&0.785&226.636&	197.786  \\
    & TCSM~\citep{li2020transformation}  &0.800&0.794&	0.681&	0.502&	7.874  &0.759&0.825&176.449&149.699 \\
    & DUW~\citep{wang2020double}    &0.815&	0.750&	0.630&	0.465&	10.913  & 0.682&0.803&195.196&153.082  \\
    & CPS~\citep{chen2021semi}    &\textbf{0.844}&	0.767&	0.644&	0.535&	7.018  & 0.723&0.828&174.738&150.858  \\
    & URPC~\citep{luo2022semi}    &0.811&	0.787&	0.677&	0.503&	8.351  & 0.492&0.643&376.769&324.264  \\
    & MC-Net~\citep{wu2022mutual}    &0.836&	0.800&	0.689&	0.548&	7.405  & 0.527&0.652&364.860&309.034  \\
    & PG-FANet-SSL  & 0.837&\textbf{0.809}&\textbf{0.698}&\textbf{0.564}&\textbf{6.641} & \textbf{0.807}&\textbf{0.869}&\textbf{136.498}&\textbf{112.232}  \\ 
    \hline
    \multirow{6}{*}{ 10\% (3/15)  }
    & PG-FANet& 0.874&0.798&	0.686&	0.580&	7.071 &0.770&0.821&192.773&169.371  \\
    & MT~\citep{tarvainen2017mean} & 0.881&	0.817&	0.708&	0.598&	6.299 &0.802&0.866&139.056&118.682  \\
    & UA-MT~\citep{yu2019uncertainty}  & 0.880& 0.818&	0.710&0.598&6.285 &0.802&0.857&	143.842&	125.867   \\
    & ICT~\citep{verma2019interpolation}  &0.879&0.817&0.709&	0.597&	6.272  &0.799&0.864&138.699&120.673  \\
    & TCSM~\citep{li2020transformation}  & 0.883&\textbf{0.819}&	0.711&	0.602&	6.181  &0.806&0.857&149.408&127.688  \\
    & DUW~\citep{wang2020double}    & 0.885&0.807&	0.697&	0.591&	6.941  & 0.790&0.853&163.726&131.580   \\
    & CPS~\citep{chen2021semi}    &0.883&	0.807&	0.696&	0.599&	\textbf{6.105}  & 0.771&0.862&150.635&123.897  \\
    & URPC~\citep{luo2022semi}    &0.876&	\textbf{0.819}&	\textbf{0.713}&	0.596&	6.179  & 0.691&0.794&206.021&174.280  \\
    & MC-Net~\citep{wu2022mutual}    &0.866&	0.817&	0.710&	0.588&	6.473  &0.543&0.690&335.600&281.984 \\
    & PG-FANet-SSL                     &\textbf{0.886}&	0.813&	0.703&\textbf{0.609}&	6.416 & \textbf{0.822}&\textbf{0.878}&\textbf{116.901}&\textbf{97.906}  \\ \hline
    \multirow{6}{*}{ 20\% (5/31) }  
    & PG-FANet&0.884&0.811&0.702&	0.602&	6.726 & 0.828&0.870&128.166&110.100   \\
    & MT~\citep{tarvainen2017mean}& 0.887&	0.822&	0.716&  	0.609& 	6.182  &0.836&0.886&116.070&\textbf{96.658}  \\
    & UA-MT~\citep{yu2019uncertainty}   & 0.887&0.826&\textbf{0.721}&	0.614&	6.000  & 0.834&0.884&118.373&99.370  \\
    & ICT~\citep{verma2019interpolation}  & 0.885&0.823&	0.717&0.611&	6.089 & 0.821&0.877&122.433&105.343  \\
    & TCSM~\citep{li2020transformation}    & 0.888&0.824&0.717&0.614&	6.078 &0.817&0.873&127.817&107.789   \\
    & DUW~\citep{wang2020double}    & 0.887&0.822&	0.715&	0.616&	6.146  & \textbf{0.837}&0.876&125.471&101.802  \\
    & CPS~\citep{chen2021semi}    &\textbf{0.891}&	0.820&	0.712&	0.625&	\textbf{5.694}  & 0.793&0.869&139.266&115.265  \\
    & URPC~\citep{luo2022semi}    &0.888&	\textbf{0.832}&	0.729&	0.621&	5.830  & 0.736&0.814&193.536&161.891 \\
    & MC-Net~\citep{wu2022mutual}    &0.872&	0.822&	0.716&	0.593&	6.484  & 0.629&0.741&283.179&236.890  \\
    & PG-FANet-SSL  & \textbf{0.891}&0.825&0.718&\textbf{0.629}&5.717 & 0.819&\textbf{0.888}&\textbf{112.694}&96.968    \\ \hline
    \multirow{6}{*}{ 50\% (14/76)}   
    & PG-FANet& 0.884&0.809&0.700&0.598&6.731 &0.836&0.887&123.998&95.639 \\
    & MT~\citep{tarvainen2017mean}& \textbf{0.891}&0.827&	0.722&	0.622&	5.900 &0.843&0.883&121.485&95.565  \\
    & UA-MT~\citep{yu2019uncertainty} & 0.885&0.829&	0.723&	0.620&	5.902 & 0.816&0.880&120.328&101.421  \\
    & ICT~\citep{verma2019interpolation}  & 0.889&0.828&	0.723&0.622&	5.865&\textbf{0.846}&0.871&143.577&112.132 \\
    & TCSM~\citep{li2020transformation}&0.888& 	0.827&	0.722&0.621&	5.913 & 0.844&0.886&	120.630&	96.050  \\
    & DUW~\citep{wang2020double}& 0.887&	0.804&	0.691&	0.594&	6.784 & 0.843&	0.886&	114.495&	99.170    \\
    & CPS~\citep{chen2021semi}    &0.886&	0.826&	0.720&	\textbf{0.630}&	\textbf{5.555}  &0.834&0.888&122.204&96.662  \\
    & URPC~\citep{luo2022semi}    &0.886&	\textbf{0.834}&	\textbf{0.732}&	0.625&	5.573  &0.725&0.830&166.954&145.954  \\
    & MC-Net~\citep{wu2022mutual}    &0.880&	0.807&	0.696&	0.595&	6.245  & 0.634&0.774&222.888&194.171  \\
   & PG-FANet-SSL & 0.890&0.826&0.720&0.626&5.777  &0.822&\textbf{0.889}&\textbf{112.926}&	\textbf{93.775} \\ \hline
    100\% (27/153)                   & PG-FANet& 0.900  & 0.839 &0.736  & 0.645 &5.420&  0.860&	0.901&	102.683&	80.181\\
    \bottomrule
  \end{tabular}
\end{table*}

\begin{table}[]
  \centering
  \caption{Experimental results on MoNuSeg and CRAG using our PG-FANet and state-of-the-art semi-supervised learning methods for nuclei/gland segmentation with different percentages of labeled data.}
  \label{table:performance_comp_using_semi_data_medical}
  \renewcommand\arraystretch{0.9}
  \renewcommand\tabcolsep{20pt}
  \begin{tabular}{cc|c|cc}
    \toprule
    \multirow{2}{*}{\textbf{\makecell[c]{Labeled \\ data }}}
    & \multirow{2}{*}{\textbf{Methods} } & \multicolumn{1}{c|}{\textbf{MoNuSeg}} & \multicolumn{2}{c}{\textbf{CRAG}} \\ \cline{3-5}
                             &                                    & \textbf{F1}        & \textbf{F1}    & \textbf{$\mathrm{\textbf{Dice}_{obj}}$} \\ \hline
    \multirow{3}{*}{20\%}
    & Self-loop~\citep{wang2020double}    & 0.771 & -&-  \\
    & $\mathrm{PRS^2}$~\citep{xie2020pairwise} & - & 0.807&0.850  \\
    &PG-FANet-SSL  & \textbf{0.891}& \textbf{0.819}&\textbf{0.888}   \\ \hline
    \multirow{3}{*}{50\%}
    & Self-loop~\citep{wang2020double}    & 0.791&-&-  \\
    & $\mathrm{PRS^2}$~\citep{xie2020pairwise}    & - & \textbf{0.823}&0.870  \\
   & PG-FANet-SSL & \textbf{0.890}&0.822&\textbf{0.889}\\
    \bottomrule
  \end{tabular}
\end{table}

\subsubsection{Nuclei segmentation} On the MoNuSeg dataset, we evaluate the performance of our proposed model on the test data in comparison with state-of-the-art methods. As shown in Table~\ref{table:monuseg_performance_comp_using_full_data}, PG-FANet outperforms all the other models in terms of all the evaluation metrics under both the same and different experimental settings. With the same experimental setting, the improvement by 1.6\% of our model is substantial compared with the second best model MedFormer~\citep{gao2022data} on AJI score, which is a key metric for nuclei segmentation. When comparing with models under different experimental settings, the improvement of our PG-FANet is also desirable on Dice and IoU scores. Due to the missing of several metrics in their original work, F1, AJI, and 95HD cannot be explicitly compared. Even with limited evaluation metrics, the experimental results demonstrate that our PG-FANet outperforms all the other models.

\subsubsection{Gland segmentation}
We evaluate the gland segmentation performance of PG-FANet with other methods including U-Net~\citep{ronneberger2015u}, DCAN~\citep{chen2016dcan}, MILD-Net~\citep{graham2019mild}, DSE~\citep{xie2019deep}, $\mathrm{PRS^2}$~\citep{xie2020pairwise}, MedFormer~\citep{gao2022data}, HARU-Net~\citep{chen2023enhancing}, and ADS$\_$UNet~\citep{yang2023ads_unet}. As shown in Table~\ref{table:crag_performance_comp_using_full_data}, PG-FANet consistently achieves the best performance in terms of F1, $\mathrm{Dice_{obj}}$, and $\mathrm{Haus_{obj}}$ among all the models.

In summary, the experimental results demonstrate that our PG-FANet with MGFE and MMFA outperforms the state-of-the-art methods and improves the fully supervised segmentation results.

\subsection{Segmentation results using limited amount of labeled data}
The major advantage of our semi-supervised framework, denoted by PG-FANet-SSL, is to use easily available unlabeled images to facilitate model training, leading to (1) less requirement on training data with annotations and (2) substantially improved segmentation performance, particularly when the number of images in the densely annotated training dataset is small.
To evaluate the performance of PG-FANet-SSL, we conduct experiments by gradually increasing the proportion of labeled data. 
We also compare our framework with recent semi-supervised models including the mean teacher model (MT)~\citep{tarvainen2017mean}, uncertainty-aware self-ensembling model (UA-MT)~\citep{yu2019uncertainty}, interpolation consistency training model (ICT)~\citep{verma2019interpolation}, transformation-consistent self-ensembling model (TCSM)~\citep{li2020transformation}, double-uncertainty weighted model (DUW)~\citep{wang2020double}, cross pseudo supervision (CPS)~\citep{chen2021semi}, uncertainty rectified pyramid consistency (URPC)~\citep{luo2022semi}, and mutual consistency learning (MC-Net)~\citep{wu2022mutual}. For URPC and MC-Net, we utilize the original backbone network instead of PG-FANet, as it is not compatible with these methods. All other SSL methods mentioned above are re-implemented using PG-FANet as backbone and conducted under the same settings. Apart from the above methods, we compare our framework with the recent semi-supervised models, i.e., Self-loop~\citep{wang2020double} and $\mathrm{PRS^2}$~\citep{xie2020pairwise}, for nuclei/gland segmentation as well. We directly copy the values from the Self-loop and $\mathrm{PRS^2}$ papers which we cannot obtain the source codes.

\subsubsection{Nuclei segmentation}
As shown in Table~\ref{table:performance_comp_using_semi_data} and Table~\ref{table:performance_comp_using_semi_data_medical}, our PG-FANet-SSL yields significant improvements over PG-FANet using the same proportion of labeled data. Specifically, with the increasing number of labeled images, our model shows 5.9\%, 2.9\%, 2.7\%, and 2.8\% improvements on AJI compared with fully supervised training. It is noted that the AJI score has an obvious improvement when the labeled data proportion increases from 5\% to 50\%, which reveals that the increasing number of the labeled data has a significant impact when there is only a small amount of data with annotations. Furthermore, our PG-FANet-SSL method achieved overall better performances on AJI over all the semi-supervised learning models in comparison. It is interesting that with 5\% labeled data, the other SSL methods gain marginal performances compared to the baseline PG-FANet. While with the increasing of labeled data, the other SSL methods show their learning ability on unlabeled data and achieve obvious improvements. We explain the finding as that our PG-FANet-SSL obtains more robust results with the inter- and intra-uncertainty regularization than the other methods.

\subsubsection{Gland segmentation}
The performance of our semi-supervised method is further demonstrated using the CRAG dataset as shown in Table~\ref{table:performance_comp_using_semi_data} and Table~\ref{table:performance_comp_using_semi_data_medical}. Compared with the fully supervised baseline, PG-FANet, our PG-FANet-SSL method significantly improves F1, $\mathrm{Dice_{obj}}$, and $\mathrm{Haus_{obj}}$ by 8.9\%/5.2\%, 9.6\%/5.7\%, and 109.632/75.872 respectively when using 5\%/10\% of labeled data. When compared with state-of-the-art methods, our PG-FANet-SSL shows significant improvement with only 5\%/10\% labeled data for training.

In summary, our first finding is that with the increasing number of labeled data, the performance of segmentation improves steadily. Second, our proposed PG-FANet-SSL achieves better performances compared to the fully supervised PG-FANet when it is trained with the same proportion of labeled data. Third, our PG-FANet-SSL maintains competitive performance when using 50\% labeled data compared with other fully supervised methods with 100\% labeled data. Fourth, PG-FANet-SSL outperforms the recent state-of-the-art semi-supervised learning methods especially with 5\%, 10\%, and 20\% labeled data for training, and this shows the effectiveness of our uncertainty modeling strategy. Last but not the least, it is interesting that the performance discrepancy between semi-supervised and supervised learning methods becomes marginal with the increase in the number of labeled data. We explain this finding as the reason that the diversity of these two datasets is limited, and a whole histopathology image may contain a certain number of labeled nuclei/gland instances. Thus, only a limited number of labeled data is needed for training to obtain state-of-the-art performance on nuclei/gland segmentations.

\subsection{Ablation studies}
   

\subsubsection{Effectiveness of MGFE and MMFA}
We perform ablation studies to evaluate the contributions of different components in our framework. We first remove all the components, degrade the two-stage network to a single-stage network (i.e., DeepLabV2~\citep{chen2017deeplab} with an extra convolutional layer), and gradually add proposed components (i.e., mask-guided, multi-scale, and multi-stage) to the model. As shown in Table~\ref{table:performance_comp_module_func}, the overall performance evaluation metric, AJI score, increases by 0.8\% on MoNuSeg when we add one more stage to DeepLabV2, which indicates that the extra backbone cannot improve the ability of the model. The AJI metric, however, increases by 1.4\% when the mask-guided feature enhancement module is introduced to the two-stage network. The MGFE module utilizes the coarse segmentation results as an enhancement and finally improves the learning ability of the model. Improvement can be observed with the application of multi-scale and multi-stage feature aggregation modules.

\begin{table}[]
    \centering
    \caption{Effectiveness analysis of different modules in PG-FANet on MoNuSeg using 100\% labeled data.}
    \label{table:performance_comp_module_func}
    \renewcommand\arraystretch{0.8}
    \renewcommand\tabcolsep{13pt}
    \begin{tabular}{cccc|ccccc}
      \toprule
     
       \textbf{\makecell[c]{ Two- \\ stage }}   &  \textbf{\makecell[c]{ Mask-\\guided}}   &\textbf{\makecell[c]{ Multi-\\scale}}   &\textbf{\makecell[c]{ Multi-\\stage}}   &\textbf{F1} & \textbf{Dice}  & \textbf{IoU}  & \textbf{AJI}  & \textbf{95HD} \\ \hline
       \ding{55} & \ding{55} &  \ding{55} &\ding{55} & 0.876&0.807&0.698&0.589&7.020   \\
       \ding{51} & \ding{55} &\ding{55} &\ding{55} & 0.882&0.809&0.701&0.597&6.826   \\
       \ding{51} & \ding{51} &\ding{55} &\ding{55} &0.886&0.821&0.715&	0.615&6.173   \\
      \ding{51} & \ding{51} &\ding{51} &\ding{55} &0.896&0.837&0.734&0.640&5.531 \\
      \ding{51} & \ding{51} &\ding{51} &\ding{51} & 0.900& 0.839 &0.736 & 0.645 &5.420   \\
      \bottomrule
    \end{tabular}
    \begin{tablenotes}
      \footnotesize
      \item \textbf{Two-stage}: Two-stage sub-networks,  \textbf{Mask-guided}: Mask-guided feature enhancement,  \textbf{Multi-scale}: Multi-scale feature aggregation,  \textbf{Multi-stage}: Multi-stage feature aggregation. 
    \end{tablenotes}
  \end{table}

   

\begin{table*}[]
    \centering
    \caption{Effectiveness analysis of regularizations in PG-FANet-SSL on MoNuSeg and CRAG using 5\% labeled data.}
    \label{table:ablation of ssl}
    \renewcommand\arraystretch{0.8}
    \renewcommand\tabcolsep{6.4pt}
    \begin{tabular}{ccc|ccccc|cccc}
      \toprule
    \multirow{2}{*}{\textbf{Inter-}}
      & \multirow{2}{*}{ \textbf{Intra-}}& \multirow{2}{*}{\textbf{Shape-}} & \multicolumn{5}{c|}{\textbf{MoNuSeg}} & \multicolumn{4}{c}{\textbf{CRAG}} \\ \cline{4-12}
      &  & & \textbf{F1}& \textbf{Dice}& \textbf{IoU}  & \textbf{AJI}  & \textbf{95HD} & \textbf{F1}    & \textbf{$\mathrm{\textbf{Dice}_{obj}}$} & \textbf{ $\mathrm{\textbf{Haus}_{obj}}$} & \textbf{$\mathrm{\textbf{95HD}_{obj}}$}  \\ \hline
      \ding{55} &  \ding{55}& \ding{55} &0.822& 0.767&0.646&0.505&8.998& 0.718&	0.773&	246.130&	208.665  \\
      \ding{51} & \ding{55} & \ding{55} &0.801&0.797&0.684&0.520&7.437 &0.796&0.858&148.080&123.161 \\
      \ding{55} & \ding{51} & \ding{55} &0.818&0.796&0.683&0.524&7.620  &0.764&0.860&140.692&119.236    \\
      \ding{55} & \ding{55} & \ding{51} &0.782&0.794&0.680&0.495&7.726 & 0.775&0.850&168.771&131.135   \\
      \ding{51} & \ding{51} & \ding{55} &0.826&0.803&0.691&0.547&7.072  &0.797&0.861&145.948&116.866 \\
      \ding{51} & \ding{51} & \ding{51} & 0.837&0.809&0.698&0.564&6.641 & 0.807&0.869&136.498&112.232 \\ 
      \bottomrule
    \end{tabular}
    \begin{tablenotes}
      \footnotesize
      \item \textbf{Inter-}: Inter-consistency regularization,  \textbf{Intra-}: Intra-consistency regularization,  \textbf{Shape-}: Shape attention weighted consistency regularization.
    \end{tablenotes}
  \end{table*}

  \begin{figure*}
    \centering
    \includegraphics[scale=0.6]{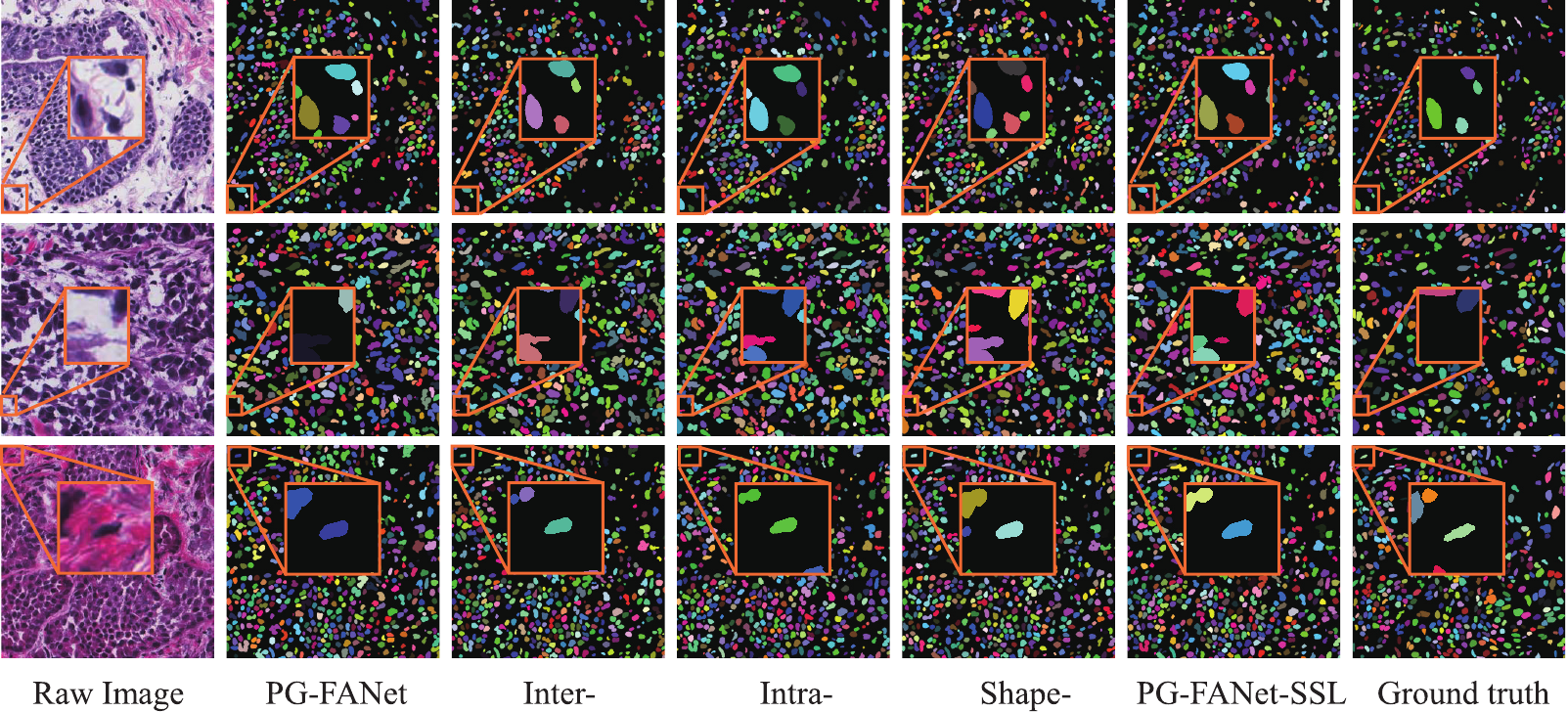}
    \caption{Segmentation results on the MoNuSeg dataset with each add-on component in PG-FANet-SSL using 5\% labeled data.    }
    \label{fig:ablation}
  \end{figure*}
\subsubsection{Effectiveness of inter- and intra-uncertainty and consistency regularization}
We use 5\% labeled NoNuSeg and CRAG to demonstrate the effectiveness of the uncertainty and consistency regularization. Table~\ref{table:ablation of ssl} presents the ablation studies of our key components. On the one hand, adding the inter-consistency regularization strategy improves AJI/$\mathrm{Dice_{obj}}$ metric by 1.5\%/8.5\% on the MoNuSeg/CRAG dataset. On the other hand, reducing intra-uncertainty increases the AJI/$\mathrm{Dice_{obj}}$ metric by 1.9\%/8.7\% as well. Furthermore, shape attention weighted consistency regularization also preserves the complete shape of segmentation in medical images. To visually illustrate the effectiveness of each component on the MoNuSeg dataset, we present the segmentation results in Figure~\ref{fig:ablation}. As depicted, the occurrences of false positive predictions significantly diminish with the aid of inter-uncertainty reduction (i.e., the third column in Figure~\ref{fig:ablation}) compared to the baseline method (i.e., the second column in Figure~\ref{fig:ablation}). This outcome underscores the profound impact of strategically reducing inter-uncertainty in achieving substantial performance advancements. Regarding intra-uncertainty reduction, the extent of false predictions is also smaller compared to those produced by the baseline, affirming the efficacy of intra-uncertainty and consistency regularization. As for the shape enhancement component, the false predictions become slightly pronounced, potentially due to the smaller size and ambiguous boundaries of nuclei compared to glands. Nevertheless, the incorporation of all the consistency regularization strategies mitigates the challenge of enforcing appearance consistency, ultimately resulting in improved performance.

In summary, Table~\ref{table:ablation of ssl} and Figure~\ref{fig:ablation} indicate that (1) inconsistency within the student model exists whilst our PG-FANet-SSL approach can model the inconsistency in a better manner, (2) without the inter-uncertainty strategy, AJI/$\mathrm{Dice_{obj}}$ continues to decrease since the inter-uncertainty strategy can dynamically leverage the uncertainty obtained by the teacher model, and (3) additional boundary information and shape enhancement benefit complete object segmentation for histopathology images.


\subsection{Qualitative results}
\subsubsection{Segmentation visualization}
\begin{figure*}
    \centering
    \includegraphics[scale=0.6]{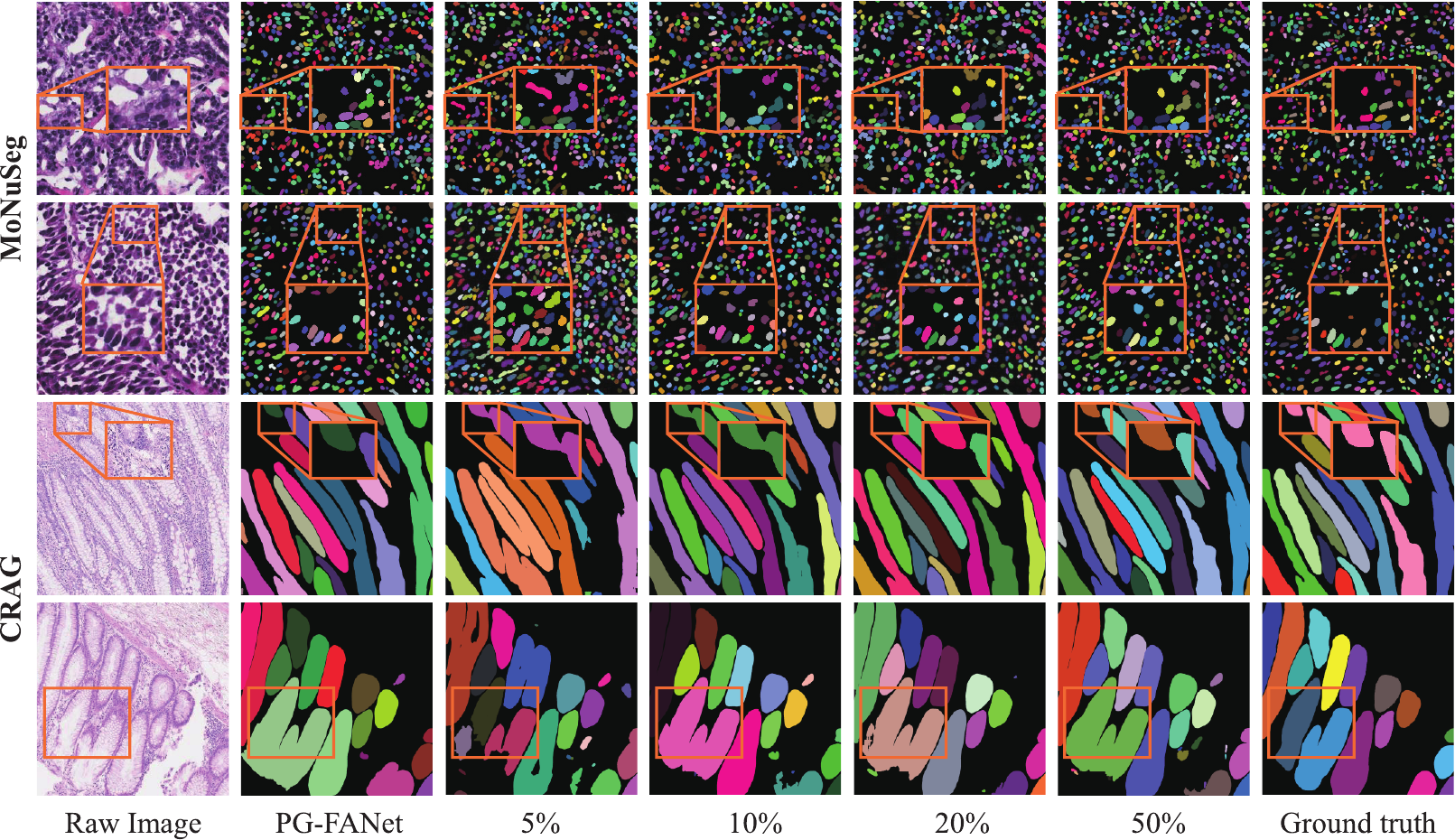}
    \caption{Segmentation results on the MoNuSeg and CRAG datasets using our fully supervised PG-FANet with 100\% labeled data and semi-supervised learning with 5\%, 10\%, 20\%, and 50\% of the labeled data.
    }
    \label{fig:stages_seg_perf}
  \end{figure*}

\begin{figure*}
  \centering
  \includegraphics[scale=0.55]{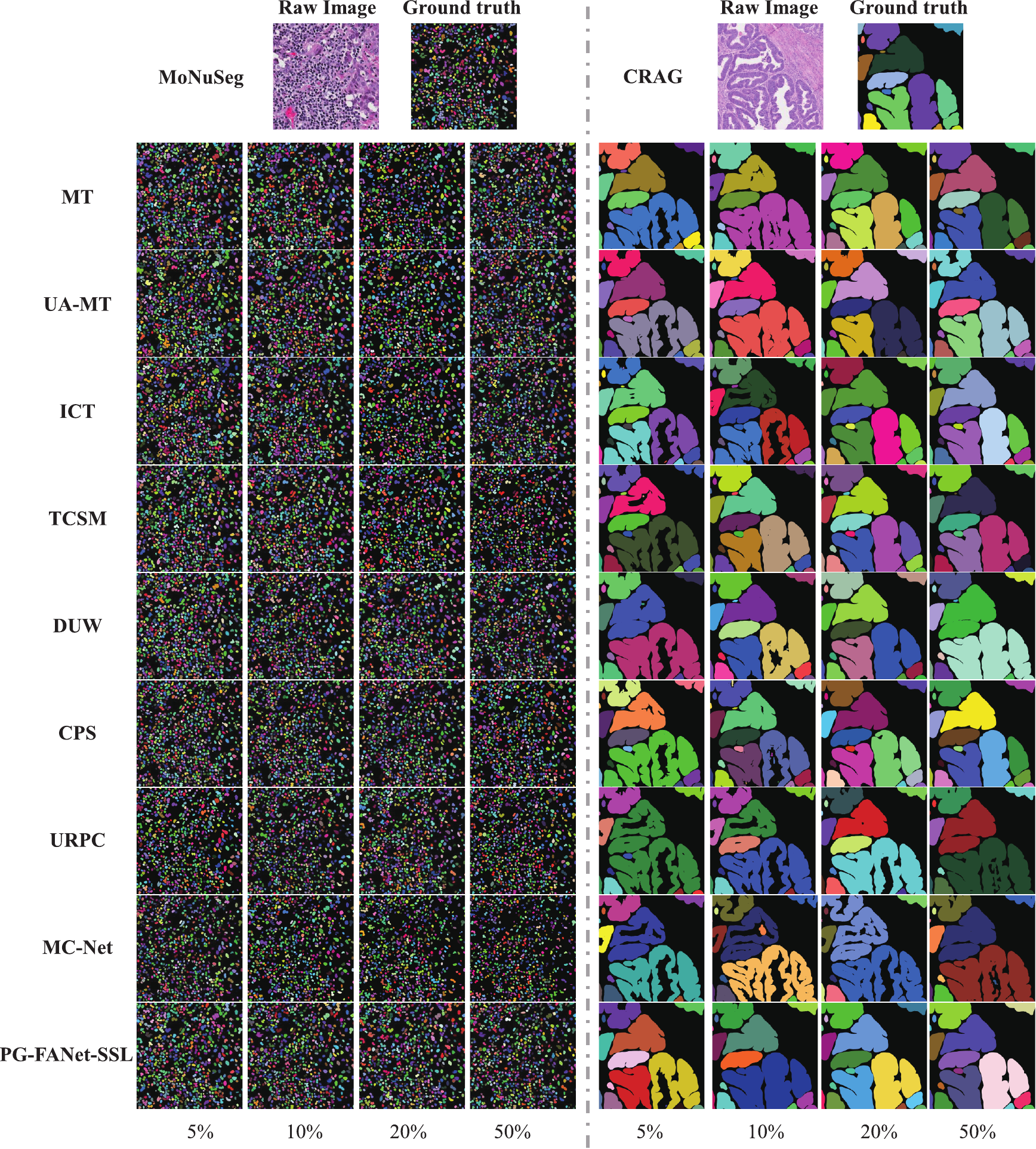}
  \caption{A representative segmentation outcome achieved with our method on the MoNuSeg and CRAG datasets, compared with results from other state-of-the-art approaches.}
  \label{fig:sota_comp}
\end{figure*}

Qualitative results on the nuclei and gland segmentations via full supervision and semi-supervision are shown in Figure~\ref{fig:stages_seg_perf}. Compared with the fully supervised PG-FANet, the PG-FANet-SSL has a competitive confident prediction near object boundaries. With the increase in the proportion of labeled data, the regions of false predictions become smaller, and the boundaries of nuclei/glands become clearer. Furthermore, we typically visualize samples generated by our PG-FANet-SSL and other state-of-the-art methods in Figure~\ref{fig:sota_comp}. As the proportion of labeled data increases, there is a noticeable enhancement in the performance of all the compared methods. This improvement can be attributed to the integration of a greater amount of valuable information into the learning process. Notably, PG-FANet-SSL stands out in terms of competitive performance when compared to all the other methods. It demonstrates clearer boundaries in the MoNuSeg and CRAG datasets, underscoring its efficacy in the context of these specific datasets.

We explain the finding as that uncertainty usually exists near object boundaries because of the subtle contrast between the foreground and background regions in histopathology images. The proposed PG-FANet-SSL framework enables the learning process to focus on such uncertainties, yielding more reliable segmentation results.

  \begin{figure*}
    \centering
    \includegraphics[scale=0.45]{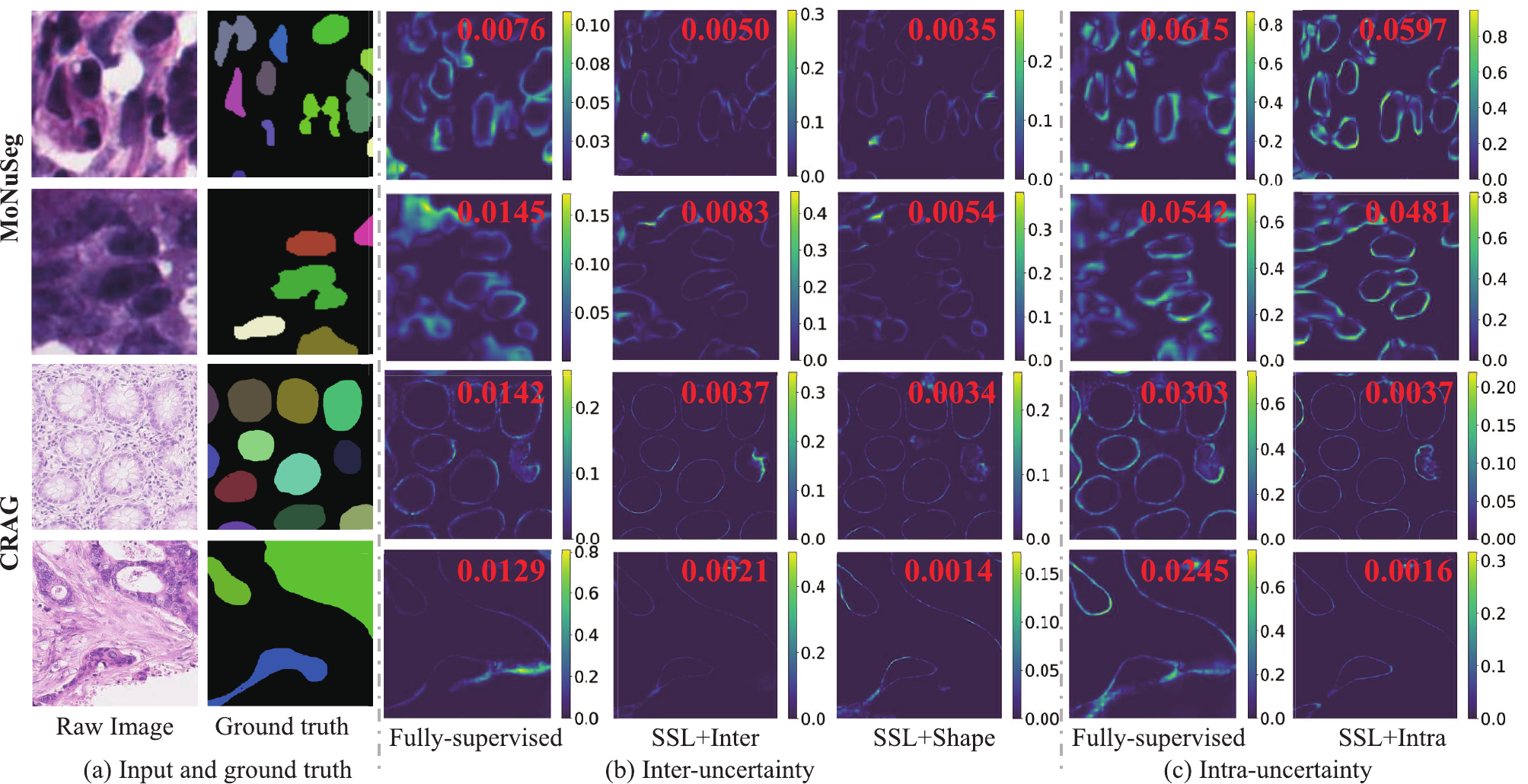}
    \caption{Histopathology image, ground truth, inter- and intra-uncertainties of our fully-supervised method, and our semi-supervised method (denoted by SSL). The average prediction variance scores in red show the inconsistencies near object boundaries. It is noted that the uncertainties between the teacher and the student models are in (b), while the uncertainties between two-stage networks are in (c).
    }
    \label{fig:vald_uncertainty_vis}
  \end{figure*}
\subsubsection{Uncertainty visualization}
We also provide the visualization results to show the inter- and intra-uncertainty differences between PG-FANet and PG-FANet-SSL trained with 5\% labeled data. As shown in Figure~\ref{fig:vald_uncertainty_vis}, we observe that the SSL with inter- or shape-attention weighted consistency regularization provides more confident boundary predictions when compared with supervised PG-FANet. Moreover, SSL with the intra-consistency regularization mechanism also reduces the prediction discrepancies within the student model. 


\subsection{Discussion}
\subsubsection{Existence of inter-uncertainties}
\begin{figure}
    \centering
    \includegraphics[scale=0.2]{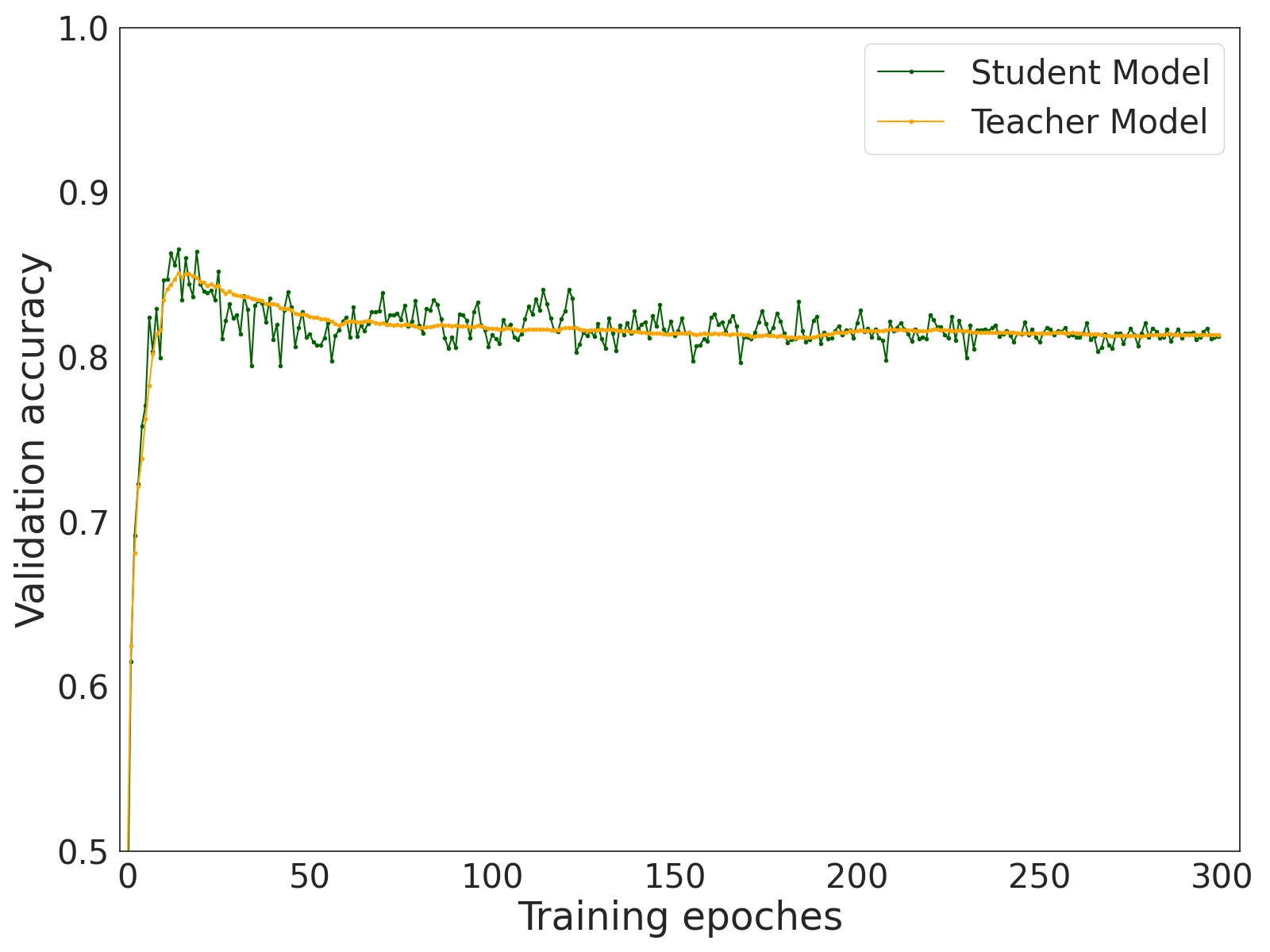}
    \caption{Segmentation accuracy on the MoNuSeg validation dataset with 5\% of the labeled data.
    }
    \label{fig:train_validation_eval}
  \end{figure}

To explore the learning ability of the teacher model and the student model, we adopt the mean-teacher architecture and train the baseline model on 5\% labeled data and 95\% unlabeled data. The performances are depicted in Figure~\ref{fig:train_validation_eval} on the validation dataset. Our primary finding is that the teacher model gains desirable stability when compared with the student model. The teacher model, however, may not always generate better results than the student model, especially at the beginning of the training process. We explain this experimental finding by the reason that the weights of the teacher mode are updated by the EMA weights of the student model which demonstrates the existence of the inter-uncertainties. With the reduction of the inter-uncertainties and prediction discrepancy, the representation ability of the model could be improved.

\subsubsection{Existence of the intra-uncertainties}
\begin{figure}
    \centering
    \includegraphics[scale=0.6]{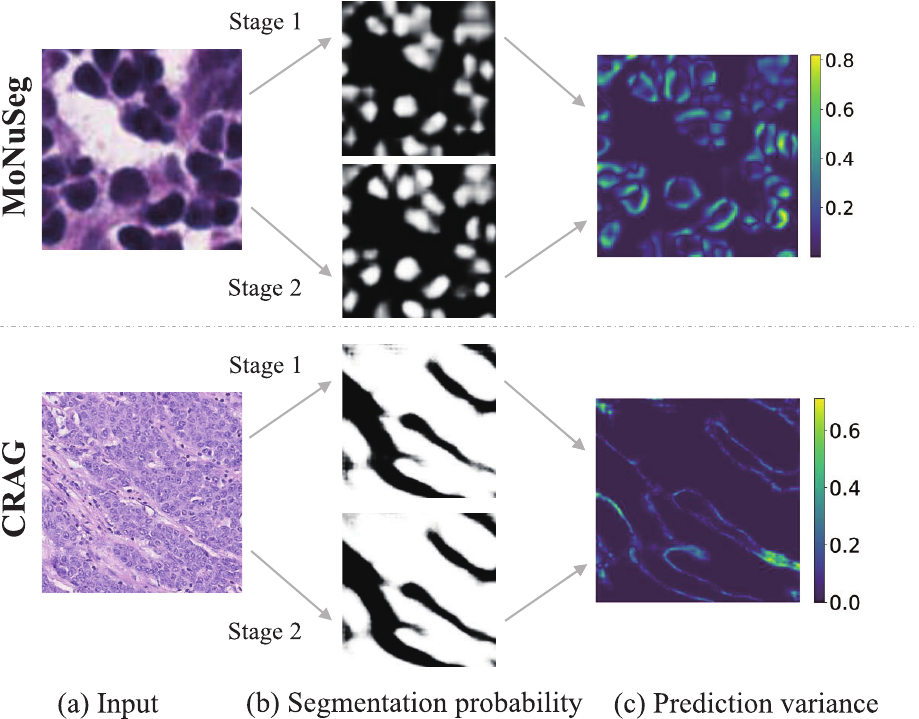}
    \caption{Illustration of the intra-uncertainty within the student model. The segmentation probability maps of input (a) between two-stage networks are in (b). In (c), brighter colors indicate higher variance. Object boundaries are with higher uncertainties.
    }
    \label{fig:uncertainty_calcu}
  \end{figure}

To further show the intra-uncertainties, we depict the two-stage prediction variances in Figure~\ref{fig:uncertainty_calcu}. The reason for such a prediction discrepancy is that different receptive fields introduce inconsistencies. As shown in Figure~\ref{fig:overview of architecture}, stage 1 is located at the relatively shallow layer, while stage 2 learns from deeper layers. The hierarchical architecture and different receptive fields increase the intra-uncertainties, which leads to prediction differences. Based on this finding, we enforce an invariance for predictions of the two stages over small perturbations applied to the hidden features. As a result, the learned model will be robust to such intra-uncertainties.

  \begin{table}[]
    \centering
    \caption{Effectiveness analysis of loss functions on MoNuSeg using 100\% labeled data.}
    \label{table:performance_comp_loss_func}
    \renewcommand\arraystretch{0.8}
    \renewcommand\tabcolsep{13.4pt}
    \begin{tabular}{c|ccccc}
      \toprule
     
       \textbf{Methods} & \textbf{F1}    & \textbf{Dice}     & \textbf{IoU}  & \textbf{AJI}& \textbf{95HD} \\ \hline
      $L_{\mathrm{ce}}$                     & 0.887&0.824&0.715 &0.625&5.456 \\
      $L_{\mathrm{ce}}$+$L_{\mathrm{Dice}}$ & 0.887&	0.827&	0.720&0.628&5.336\\
      $L_{\mathrm{ce}}$+$L_{\mathrm{vcc}}$  & 0.891&0.834&0.731&0.635&	5.665 \\
      $L_{\mathrm{ce}}$+$L_{\mathrm{Dice}}$+$L_{\mathrm{vcc}}$  & 0.900 & 0.839&0.736    & 0.645 &5.420 \\
      \bottomrule
    \end{tabular}
  \end{table}

\subsubsection{Effectiveness of loss functions}
To demonstrate the performance gain obtained from the proposed segmentation loss (i.e., $L_{\mathrm{seg}}$), we also attempt to train PG-FANet with different loss functions (i.e., $L_{\mathrm{ce}}$, $L_{\mathrm{Dice}}$, and $L_{\mathrm{vcc}}$) and their combinations. The results in Table~\ref{table:performance_comp_loss_func} reveal that, although using $L_{\mathrm{ce}}$ alone results in desirable performances on the MoNuSeg dataset, adding $L_{\mathrm{Dice}}$ and $L_{\mathrm{vcc}}$ individually to the segmentation loss improves the performance continuously. Meanwhile, the superior performance of our $L_{\mathrm{seg}}$ confirms the effectiveness of the combination of $L_{\mathrm{ce}}$, $L_{\mathrm{Dice}}$, and $L_{\mathrm{vcc}}$ loss functions, which have constraints on each individual instance.

\begin{table}[]
  \centering
  \caption{Flexibility analysis of single-stage PG-FANet on MoNuSeg using 5\% labeled data.}
  \label{table:performance_comp_degrade}
  \renewcommand\arraystretch{0.9}
  \renewcommand\tabcolsep{12.4pt}
  \begin{tabular}{c|ccccc}
    \toprule
     \textbf{Methods} & \textbf{F1}    & \textbf{Dice}     & \textbf{IoU}  & \textbf{AJI}& \textbf{95HD} \\ \hline
    PG-FANet                     & 0.822& 0.767&	0.646&0.505&8.998    \\
    DeepLabV2~\citep{chen2017deeplab}    & 0.797&0.717&0.590&0.440&12.140    \\
    MT~\citep{tarvainen2017mean} & 0.782&0.771&0.654&0.474&9.087  \\
    MT~\citep{tarvainen2017mean}+Inter  &0.812&0.778&0.662&0.509&8.493 \\
    MT~\citep{tarvainen2017mean}+Shape  &0.784&0.773&0.656&0.477&8.978  \\
    MT~\citep{tarvainen2017mean}+Inter+Shape  & 0.819&0.781&0.665&0.517&8.289  \\
    \bottomrule
  \end{tabular}
\end{table}

\subsubsection{The flexibility of the proposed algorithm}
To demonstrate the flexibility of the proposed consistency regularization algorithm, we also degrade the two-stage network. In Table~\ref{table:performance_comp_degrade}, PG-FANet and DeepLabV2~\citep{chen2017deeplab} denote fully supervised two-stage PG-FANet and degraded PG-FANet trained with only 5\% labeled data, respectively. Furthermore, we adopt DeepLabV2~\citep{chen2017deeplab} as our supervised training backbone in the MT~\citep{tarvainen2017mean} architecture, and gradually add proposed components. It is noted that without the two-stage architecture, intra-consistency regularization is not applicable. As shown in Table~\ref{table:performance_comp_degrade}, when the two-stage architecture is not adopted, the performance dramatically decreases from 50.5\% to 44.0\% on AJI. With the proposed consistency constraint and regularization, our model is substantially more accurate on nuclei segmentation.
This experimental results reveal that (1) our two-stage network with MGFE and MMFA substantially improves the segmentation performance, (2) the proposed inter-consistency and shape attention weighted consistency regularization strategy also functions well on degraded models, and (3) the proposed module and learning strategy are simple yet flexible for extending to other models.


\section{Conclusion}
\label{sec:conclusion}
In this paper, we first propose a novel pseudo-mask guided feature aggregation network to leverage multi-scale and multi-stage features for histopathology image segmentation. Our semi-supervised learning framework models uncertainties in the mean teacher architecture by a novel inter- and intra-uncertainty and consistency regularization strategy. Comprehensive experimental results show that our semi-supervised segmentation model achieves competitive results for nuclei and gland segmentations using partially labeled data when compared with fully- or semi-supervised models. One limitation of our work pertains to the large number of parameters, primarily attributable to the selected backbone. Our future work includes the parameter reduction and extension to other 3D image segmentation tasks.

\bibliography{refs}

\begin{thebibliography}{86}
\expandafter\ifx\csname natexlab\endcsname\relax\def\natexlab#1{#1}\fi
\providecommand{\url}[1]{\texttt{#1}}
\providecommand{\href}[2]{#2}
\providecommand{\path}[1]{#1}
\providecommand{\DOIprefix}{doi:}
\providecommand{\ArXivprefix}{arXiv:}
\providecommand{\URLprefix}{URL: }
\providecommand{\Pubmedprefix}{pmid:}
\providecommand{\doi}[1]{\href{http://dx.doi.org/#1}{\path{#1}}}
\providecommand{\Pubmed}[1]{\href{pmid:#1}{\path{#1}}}
\providecommand{\bibinfo}[2]{#2}
\ifx\xfnm\relax \def\xfnm[#1]{\unskip,\space#1}\fi
\bibitem[{Alom et~al.(2018)Alom, Yakopcic, Taha \& Asari}]{alom2018nuclei}
\bibinfo{author}{Alom, M.~Z.}, \bibinfo{author}{Yakopcic, C.},
  \bibinfo{author}{Taha, T.~M.}, \& \bibinfo{author}{Asari, V.~K.}
  (\bibinfo{year}{2018}).
\newblock \bibinfo{title}{{Nuclei segmentation with recurrent residual
  convolutional neural networks based U-Net (R2U-Net)}}.
\newblock In {\it \bibinfo{booktitle}{NAECON 2018-IEEE National Aerospace and
  Electronics Conference}\/} (pp. \bibinfo{pages}{228--233}).
\newblock \bibinfo{organization}{IEEE}.
\bibitem[{Awan et~al.(2017)Awan, Sirinukunwattana, Epstein, Jefferyes, Qidwai,
  Aftab, Mujeeb, Snead \& Rajpoot}]{awan2017glandular}
\bibinfo{author}{Awan, R.}, \bibinfo{author}{Sirinukunwattana, K.},
  \bibinfo{author}{Epstein, D.}, \bibinfo{author}{Jefferyes, S.},
  \bibinfo{author}{Qidwai, U.}, \bibinfo{author}{Aftab, Z.},
  \bibinfo{author}{Mujeeb, I.}, \bibinfo{author}{Snead, D.}, \&
  \bibinfo{author}{Rajpoot, N.} (\bibinfo{year}{2017}).
\newblock \bibinfo{title}{Glandular morphometrics for objective grading of
  colorectal adenocarcinoma histology images}.
\newblock {\it \bibinfo{journal}{Scientific Reports}\/},  {\it
  \bibinfo{volume}{7}\/}, \bibinfo{pages}{1--12}.
\bibitem[{Bai et~al.(2023)Bai, Chen, Li, Shen \& Wang}]{bai2023bidirectional}
\bibinfo{author}{Bai, Y.}, \bibinfo{author}{Chen, D.}, \bibinfo{author}{Li,
  Q.}, \bibinfo{author}{Shen, W.}, \& \bibinfo{author}{Wang, Y.}
  (\bibinfo{year}{2023}).
\newblock \bibinfo{title}{{Bidirectional Copy-Paste for Semi-Supervised Medical
  Image Segmentation}}.
\newblock In {\it \bibinfo{booktitle}{Proceedings of the IEEE/CVF Conference on
  Computer Vision and Pattern Recognition}\/} (pp.
  \bibinfo{pages}{11514--11524}).
\bibitem[{Basak \& Yin(2023)}]{basak2023pseudo}
\bibinfo{author}{Basak, H.}, \& \bibinfo{author}{Yin, Z.}
  (\bibinfo{year}{2023}).
\newblock \bibinfo{title}{Pseudo-label guided contrastive learning for
  semi-supervised medical image segmentation}.
\newblock In {\it \bibinfo{booktitle}{Proceedings of the IEEE/CVF Conference on
  Computer Vision and Pattern Recognition}\/} (pp.
  \bibinfo{pages}{19786--19797}).
\bibitem[{Cao et~al.(2022)Cao, Chen, Li, Peng, Zhou, Cheng, Liu \&
  Shen}]{cao2022auto}
\bibinfo{author}{Cao, X.}, \bibinfo{author}{Chen, H.}, \bibinfo{author}{Li,
  Y.}, \bibinfo{author}{Peng, Y.}, \bibinfo{author}{Zhou, Y.},
  \bibinfo{author}{Cheng, L.}, \bibinfo{author}{Liu, T.}, \&
  \bibinfo{author}{Shen, D.} (\bibinfo{year}{2022}).
\newblock \bibinfo{title}{{Auto-DenseUNet: Searchable neural network
  architecture for mass segmentation in 3D automated breast ultrasound}}.
\newblock {\it \bibinfo{journal}{Medical Image Analysis}\/},  {\it
  \bibinfo{volume}{82}\/}, \bibinfo{pages}{102589}.
\bibitem[{Chaitanya et~al.(2023)Chaitanya, Erdil, Karani \&
  Konukoglu}]{chaitanya2023local}
\bibinfo{author}{Chaitanya, K.}, \bibinfo{author}{Erdil, E.},
  \bibinfo{author}{Karani, N.}, \& \bibinfo{author}{Konukoglu, E.}
  (\bibinfo{year}{2023}).
\newblock \bibinfo{title}{Local contrastive loss with pseudo-label based
  self-training for semi-supervised medical image segmentation}.
\newblock {\it \bibinfo{journal}{Medical Image Analysis}\/},  {\it
  \bibinfo{volume}{87}\/}, \bibinfo{pages}{102792}.
\bibitem[{Chaurasia \& Culurciello(2017)}]{chaurasia2017linknet}
\bibinfo{author}{Chaurasia, A.}, \& \bibinfo{author}{Culurciello, E.}
  (\bibinfo{year}{2017}).
\newblock \bibinfo{title}{{LinkNet: Exploiting encoder representations for
  efficient semantic segmentation}}.
\newblock In {\it \bibinfo{booktitle}{2017 IEEE Visual Communications and Image
  Processing (VCIP)}\/} (pp. \bibinfo{pages}{1--4}).
\newblock \bibinfo{organization}{IEEE}.
\bibitem[{Chen et~al.(2016)Chen, Qi, Yu \& Heng}]{chen2016dcan}
\bibinfo{author}{Chen, H.}, \bibinfo{author}{Qi, X.}, \bibinfo{author}{Yu, L.},
  \& \bibinfo{author}{Heng, P.-A.} (\bibinfo{year}{2016}).
\newblock \bibinfo{title}{{DCAN: deep contour-aware networks for accurate gland
  segmentation}}.
\newblock In {\it \bibinfo{booktitle}{Proceedings of the IEEE Conference on
  Computer Vision and Pattern Recognition}\/} (pp.
  \bibinfo{pages}{2487--2496}).
\bibitem[{Chen et~al.(2023)Chen, Huang, Chen, Qian \& Yu}]{chen2023enhancing}
\bibinfo{author}{Chen, J.}, \bibinfo{author}{Huang, Q.}, \bibinfo{author}{Chen,
  Y.}, \bibinfo{author}{Qian, L.}, \& \bibinfo{author}{Yu, C.}
  (\bibinfo{year}{2023}).
\newblock \bibinfo{title}{{Enhancing Nucleus Segmentation with HARU-Net: A
  Hybrid Attention Based Residual U-Blocks Network}}.
\newblock {\it \bibinfo{journal}{arXiv preprint arXiv:2308.03382}\/}, .
\bibitem[{Chen et~al.(2018)Chen, Collins, Zhu, Papandreou, Zoph, Schroff, Adam
  \& Shlens}]{chen2018searching}
\bibinfo{author}{Chen, L.-C.}, \bibinfo{author}{Collins, M.},
  \bibinfo{author}{Zhu, Y.}, \bibinfo{author}{Papandreou, G.},
  \bibinfo{author}{Zoph, B.}, \bibinfo{author}{Schroff, F.},
  \bibinfo{author}{Adam, H.}, \& \bibinfo{author}{Shlens, J.}
  (\bibinfo{year}{2018}).
\newblock \bibinfo{title}{Searching for efficient multi-scale architectures for
  dense image prediction}.
\newblock In {\it \bibinfo{booktitle}{Advances in Neural Information Processing
  Systems}\/} (pp. \bibinfo{pages}{8699--8710}).
\bibitem[{Chen et~al.(2017)Chen, Papandreou, Kokkinos, Murphy \&
  Yuille}]{chen2017deeplab}
\bibinfo{author}{Chen, L.-C.}, \bibinfo{author}{Papandreou, G.},
  \bibinfo{author}{Kokkinos, I.}, \bibinfo{author}{Murphy, K.}, \&
  \bibinfo{author}{Yuille, A.~L.} (\bibinfo{year}{2017}).
\newblock \bibinfo{title}{{DeepLab: Semantic image segmentation with deep
  convolutional nets, atrous convolution, and fully connected CRFs}}.
\newblock {\it \bibinfo{journal}{IEEE Transactions on Pattern Analysis and
  Machine Intelligence}\/},  {\it \bibinfo{volume}{40}\/},
  \bibinfo{pages}{834--848}.
\bibitem[{Chen et~al.(2021)Chen, Yuan, Zeng \& Wang}]{chen2021semi}
\bibinfo{author}{Chen, X.}, \bibinfo{author}{Yuan, Y.}, \bibinfo{author}{Zeng,
  G.}, \& \bibinfo{author}{Wang, J.} (\bibinfo{year}{2021}).
\newblock \bibinfo{title}{Semi-supervised semantic segmentation with cross
  pseudo supervision}.
\newblock In {\it \bibinfo{booktitle}{Proceedings of the IEEE/CVF Conference on
  Computer Vision and Pattern Recognition}\/} (pp.
  \bibinfo{pages}{2613--2622}).
\bibitem[{Cui et~al.(2022)Cui, Li, Zhang, Sui, Cao, Hesham \&
  Zou}]{cui2022deepmc}
\bibinfo{author}{Cui, F.}, \bibinfo{author}{Li, S.}, \bibinfo{author}{Zhang,
  Z.}, \bibinfo{author}{Sui, M.}, \bibinfo{author}{Cao, C.},
  \bibinfo{author}{Hesham, A. E.-L.}, \& \bibinfo{author}{Zou, Q.}
  (\bibinfo{year}{2022}).
\newblock \bibinfo{title}{{DeepMC-iNABP: Deep learning for multiclass
  identification and classification of nucleic acid-binding proteins}}.
\newblock {\it \bibinfo{journal}{Computational and Structural Biotechnology
  Journal}\/},  {\it \bibinfo{volume}{20}\/}, \bibinfo{pages}{2020--2028}.
\bibitem[{Deng et~al.(2009)Deng, Dong, Socher, Li, Li \&
  Fei-Fei}]{deng2009imagenet}
\bibinfo{author}{Deng, J.}, \bibinfo{author}{Dong, W.},
  \bibinfo{author}{Socher, R.}, \bibinfo{author}{Li, L.-J.},
  \bibinfo{author}{Li, K.}, \& \bibinfo{author}{Fei-Fei, L.}
  (\bibinfo{year}{2009}).
\newblock \bibinfo{title}{{ImageNet: A large-scale hierarchical image
  database}}.
\newblock In {\it \bibinfo{booktitle}{2009 IEEE Conference on Computer Vision
  and Pattern Recognition}\/} (pp. \bibinfo{pages}{248--255}).
\newblock \bibinfo{organization}{IEEE}.
\bibitem[{Dou et~al.(2019)Dou, Ouyang, Chen, Chen, Glocker, Zhuang \&
  Heng}]{PnPAdaNet}
\bibinfo{author}{Dou, Q.}, \bibinfo{author}{Ouyang, C.}, \bibinfo{author}{Chen,
  C.}, \bibinfo{author}{Chen, H.}, \bibinfo{author}{Glocker, B.},
  \bibinfo{author}{Zhuang, X.}, \& \bibinfo{author}{Heng, P.-A.}
  (\bibinfo{year}{2019}).
\newblock \bibinfo{title}{{PnP-AdaNet: Plug-and-Play Adversarial Domain
  Adaptation Network at Unpaired Cross-Modality Cardiac Segmentation}}.
\newblock {\it \bibinfo{journal}{IEEE Access}\/},  {\it \bibinfo{volume}{7}\/},
  \bibinfo{pages}{99065--99076}. \DOIprefix\doi{10.1109/ACCESS.2019.2929258}.
\bibitem[{Gal \& Ghahramani(2016)}]{gal2016dropout}
\bibinfo{author}{Gal, Y.}, \& \bibinfo{author}{Ghahramani, Z.}
  (\bibinfo{year}{2016}).
\newblock \bibinfo{title}{{Dropout as a Bayesian approximation: Representing
  model uncertainty in deep learning}}.
\newblock In {\it \bibinfo{booktitle}{International Conference on Machine
  Learning}\/} (pp. \bibinfo{pages}{1050--1059}).
\bibitem[{Gao et~al.(2022)Gao, Zhou, Liu, Yan, Zhang \& Metaxas}]{gao2022data}
\bibinfo{author}{Gao, Y.}, \bibinfo{author}{Zhou, M.}, \bibinfo{author}{Liu,
  D.}, \bibinfo{author}{Yan, Z.}, \bibinfo{author}{Zhang, S.}, \&
  \bibinfo{author}{Metaxas, D.~N.} (\bibinfo{year}{2022}).
\newblock \bibinfo{title}{A data-scalable transformer for medical image
  segmentation: architecture, model efficiency, and benchmark}.
\newblock {\it \bibinfo{journal}{arXiv preprint arXiv:2203.00131}\/}, .
\bibitem[{Graham et~al.(2019)Graham, Chen, Gamper, Dou, Heng, Snead, Tsang \&
  Rajpoot}]{graham2019mild}
\bibinfo{author}{Graham, S.}, \bibinfo{author}{Chen, H.},
  \bibinfo{author}{Gamper, J.}, \bibinfo{author}{Dou, Q.},
  \bibinfo{author}{Heng, P.-A.}, \bibinfo{author}{Snead, D.},
  \bibinfo{author}{Tsang, Y.~W.}, \& \bibinfo{author}{Rajpoot, N.}
  (\bibinfo{year}{2019}).
\newblock \bibinfo{title}{{MILD-Net: minimal information loss dilated network
  for gland instance segmentation in colon histology images}}.
\newblock {\it \bibinfo{journal}{Medical Image Analysis}\/},  {\it
  \bibinfo{volume}{52}\/}, \bibinfo{pages}{199--211}.
\bibitem[{Gu et~al.(2022)Gu, Zhang, Wang, Lei, Song, Zhang, Li \&
  Zhang}]{gu2022contrastive}
\bibinfo{author}{Gu, R.}, \bibinfo{author}{Zhang, J.}, \bibinfo{author}{Wang,
  G.}, \bibinfo{author}{Lei, W.}, \bibinfo{author}{Song, T.},
  \bibinfo{author}{Zhang, X.}, \bibinfo{author}{Li, K.}, \&
  \bibinfo{author}{Zhang, S.} (\bibinfo{year}{2022}).
\newblock \bibinfo{title}{Contrastive semi-supervised learning for domain
  adaptive segmentation across similar anatomical structures}.
\newblock {\it \bibinfo{journal}{IEEE Transactions on Medical Imaging}\/},
  {\it \bibinfo{volume}{42}\/}, \bibinfo{pages}{245--256}.
\bibitem[{Gustafsson et~al.(2020)Gustafsson, Danelljan \&
  Schon}]{gustafsson2020evaluating}
\bibinfo{author}{Gustafsson, F.~K.}, \bibinfo{author}{Danelljan, M.}, \&
  \bibinfo{author}{Schon, T.~B.} (\bibinfo{year}{2020}).
\newblock \bibinfo{title}{{Evaluating scalable Bayesian deep learning methods
  for robust computer vision}}.
\newblock In {\it \bibinfo{booktitle}{Proceedings of the IEEE/CVF Conference on
  Computer Vision and Pattern Recognition Workshops}\/} (pp.
  \bibinfo{pages}{318--319}).
\bibitem[{He et~al.(2015)He, Zhang, Ren \& Sun}]{he2015delving}
\bibinfo{author}{He, K.}, \bibinfo{author}{Zhang, X.}, \bibinfo{author}{Ren,
  S.}, \& \bibinfo{author}{Sun, J.} (\bibinfo{year}{2015}).
\newblock \bibinfo{title}{Delving deep into rectifiers: Surpassing human-level
  performance on imagenet classification}.
\newblock In {\it \bibinfo{booktitle}{Proceedings of the IEEE International
  Conference on Computer Vision}\/} (pp. \bibinfo{pages}{1026--1034}).
\bibitem[{He et~al.(2016)He, Zhang, Ren \& Sun}]{he2016deep}
\bibinfo{author}{He, K.}, \bibinfo{author}{Zhang, X.}, \bibinfo{author}{Ren,
  S.}, \& \bibinfo{author}{Sun, J.} (\bibinfo{year}{2016}).
\newblock \bibinfo{title}{Deep residual learning for image recognition}.
\newblock In {\it \bibinfo{booktitle}{Proceedings of the IEEE Conference on
  Computer Vision and Pattern Recognition}\/} (pp. \bibinfo{pages}{770--778}).
\bibitem[{Ibtehaz \& Rahman(2020)}]{ibtehaz2020multiresunet}
\bibinfo{author}{Ibtehaz, N.}, \& \bibinfo{author}{Rahman, M.~S.}
  (\bibinfo{year}{2020}).
\newblock \bibinfo{title}{{MultiResUNet: Rethinking the U-Net architecture for
  multimodal biomedical image segmentation}}.
\newblock {\it \bibinfo{journal}{Neural Networks}\/},  {\it
  \bibinfo{volume}{121}\/}, \bibinfo{pages}{74--87}.
\bibitem[{Ioffe \& Szegedy(2015)}]{ioffe2015batch}
\bibinfo{author}{Ioffe, S.}, \& \bibinfo{author}{Szegedy, C.}
  (\bibinfo{year}{2015}).
\newblock \bibinfo{title}{{Batch normalization: Accelerating deep network
  training by reducing internal covariate shift}}.
\newblock {\it \bibinfo{journal}{arXiv preprint arXiv:1502.03167}\/}, .
\bibitem[{Javed et~al.(2020)Javed, Mahmood, Fraz, Koohbanani, Benes, Tsang,
  Hewitt, Epstein, Snead \& Rajpoot}]{javed2020cellular}
\bibinfo{author}{Javed, S.}, \bibinfo{author}{Mahmood, A.},
  \bibinfo{author}{Fraz, M.~M.}, \bibinfo{author}{Koohbanani, N.~A.},
  \bibinfo{author}{Benes, K.}, \bibinfo{author}{Tsang, Y.-W.},
  \bibinfo{author}{Hewitt, K.}, \bibinfo{author}{Epstein, D.},
  \bibinfo{author}{Snead, D.}, \& \bibinfo{author}{Rajpoot, N.}
  (\bibinfo{year}{2020}).
\newblock \bibinfo{title}{Cellular community detection for tissue phenotyping
  in colorectal cancer histology images}.
\newblock {\it \bibinfo{journal}{Medical Image Analysis}\/},  {\it
  \bibinfo{volume}{63}\/}, \bibinfo{pages}{101696}.
\bibitem[{Ji et~al.(2020)Ji, Zhang, Li, Ren, Zhang \& Luo}]{ji2020uxnet}
\bibinfo{author}{Ji, Y.}, \bibinfo{author}{Zhang, R.}, \bibinfo{author}{Li,
  Z.}, \bibinfo{author}{Ren, J.}, \bibinfo{author}{Zhang, S.}, \&
  \bibinfo{author}{Luo, P.} (\bibinfo{year}{2020}).
\newblock \bibinfo{title}{{UXNet: Searching Multi-level Feature Aggregation for
  3D Medical Image Segmentation}}.
\newblock In {\it \bibinfo{booktitle}{International Conference on Medical Image
  Computing and Computer Assisted Intervention}\/} (pp.
  \bibinfo{pages}{346--356}).
\newblock \bibinfo{organization}{Springer}.
\bibitem[{Jiang et~al.(2023)Jiang, Zhang, Zhou, Wang \& Chen}]{jiang2023donet}
\bibinfo{author}{Jiang, H.}, \bibinfo{author}{Zhang, R.},
  \bibinfo{author}{Zhou, Y.}, \bibinfo{author}{Wang, Y.}, \&
  \bibinfo{author}{Chen, H.} (\bibinfo{year}{2023}).
\newblock \bibinfo{title}{{DoNet: Deep De-overlapping Network for Cytology
  Instance Segmentation}}.
\newblock In {\it \bibinfo{booktitle}{Proceedings of the IEEE/CVF Conference on
  Computer Vision and Pattern Recognition}\/} (pp.
  \bibinfo{pages}{15641--15650}).
\bibitem[{Jin et~al.(2022)Jin, Cui, Sun, Zheng, Wei, Fang, Meng \&
  Su}]{jin2022semi}
\bibinfo{author}{Jin, Q.}, \bibinfo{author}{Cui, H.}, \bibinfo{author}{Sun,
  C.}, \bibinfo{author}{Zheng, J.}, \bibinfo{author}{Wei, L.},
  \bibinfo{author}{Fang, Z.}, \bibinfo{author}{Meng, Z.}, \&
  \bibinfo{author}{Su, R.} (\bibinfo{year}{2022}).
\newblock \bibinfo{title}{Semi-supervised histological image segmentation via
  hierarchical consistency enforcement}.
\newblock In {\it \bibinfo{booktitle}{International Conference on Medical Image
  Computing and Computer-Assisted Intervention}\/} (pp.
  \bibinfo{pages}{3--13}).
\newblock \bibinfo{organization}{Springer}.
\bibitem[{Kendall \& Gal(2017)}]{kendall2017uncertainties}
\bibinfo{author}{Kendall, A.}, \& \bibinfo{author}{Gal, Y.}
  (\bibinfo{year}{2017}).
\newblock \bibinfo{title}{{What uncertainties do we need in Bayesian deep
  learning for computer vision?}}
\newblock In {\it \bibinfo{booktitle}{Advances in Neural Information Processing
  Systems}\/} (pp. \bibinfo{pages}{5574--5584}).
\bibitem[{Kumar et~al.(2019)Kumar, Verma, Anand, Zhou, Onder, Tsougenis, Chen,
  Heng, Li, Hu et~al.}]{kumar2019multi}
\bibinfo{author}{Kumar, N.}, \bibinfo{author}{Verma, R.},
  \bibinfo{author}{Anand, D.}, \bibinfo{author}{Zhou, Y.},
  \bibinfo{author}{Onder, O.~F.}, \bibinfo{author}{Tsougenis, E.},
  \bibinfo{author}{Chen, H.}, \bibinfo{author}{Heng, P.-A.},
  \bibinfo{author}{Li, J.}, \bibinfo{author}{Hu, Z.} et~al.
  (\bibinfo{year}{2019}).
\newblock \bibinfo{title}{A multi-organ nucleus segmentation challenge}.
\newblock {\it \bibinfo{journal}{IEEE Transactions on Medical Imaging}\/},
  {\it \bibinfo{volume}{39}\/}, \bibinfo{pages}{1380--1391}.
\bibitem[{Kwon et~al.(2020)Kwon, Won, Kim \& Paik}]{kwon2020uncertainty}
\bibinfo{author}{Kwon, Y.}, \bibinfo{author}{Won, J.-H.}, \bibinfo{author}{Kim,
  B.~J.}, \& \bibinfo{author}{Paik, M.~C.} (\bibinfo{year}{2020}).
\newblock \bibinfo{title}{{Uncertainty quantification using Bayesian neural
  networks in classification: Application to biomedical image segmentation}}.
\newblock {\it \bibinfo{journal}{Computational Statistics \& Data Analysis}\/},
   {\it \bibinfo{volume}{142}\/}, \bibinfo{pages}{106816}.
\bibitem[{Lei et~al.(2022)Lei, Zhang, Du, Wang, Wan \& Nandi}]{lei2022semi}
\bibinfo{author}{Lei, T.}, \bibinfo{author}{Zhang, D.}, \bibinfo{author}{Du,
  X.}, \bibinfo{author}{Wang, X.}, \bibinfo{author}{Wan, Y.}, \&
  \bibinfo{author}{Nandi, A.~K.} (\bibinfo{year}{2022}).
\newblock \bibinfo{title}{Semi-supervised medical image segmentation using
  adversarial consistency learning and dynamic convolution network}.
\newblock {\it \bibinfo{journal}{IEEE Transactions on Medical Imaging}\/}, .
\bibitem[{Li et~al.(2019)Li, Xiong, Fan \& Sun}]{li2019dfanet}
\bibinfo{author}{Li, H.}, \bibinfo{author}{Xiong, P.}, \bibinfo{author}{Fan,
  H.}, \& \bibinfo{author}{Sun, J.} (\bibinfo{year}{2019}).
\newblock \bibinfo{title}{{DFANet: Deep feature aggregation for real-time
  semantic segmentation}}.
\newblock In {\it \bibinfo{booktitle}{Proceedings of the IEEE Conference on
  Computer Vision and Pattern Recognition}\/} (pp.
  \bibinfo{pages}{9522--9531}).
\bibitem[{Li et~al.(2020{\natexlab{a}})Li, Wang, Yu \& Heng}]{li2020dual}
\bibinfo{author}{Li, K.}, \bibinfo{author}{Wang, S.}, \bibinfo{author}{Yu, L.},
  \& \bibinfo{author}{Heng, P.-A.} (\bibinfo{year}{2020}{\natexlab{a}}).
\newblock \bibinfo{title}{Dual-teacher: Integrating intra-domain and
  inter-domain teachers for annotation-efficient cardiac segmentation}.
\newblock In {\it \bibinfo{booktitle}{International Conference on Medical Image
  Computing and Computer Assisted Intervention}\/} (pp.
  \bibinfo{pages}{418--427}).
\newblock \bibinfo{organization}{Springer}.
\bibitem[{Li et~al.(2020{\natexlab{b}})Li, Yu, Chen, Fu, Xing \&
  Heng}]{li2020transformation}
\bibinfo{author}{Li, X.}, \bibinfo{author}{Yu, L.}, \bibinfo{author}{Chen, H.},
  \bibinfo{author}{Fu, C.-W.}, \bibinfo{author}{Xing, L.}, \&
  \bibinfo{author}{Heng, P.-A.} (\bibinfo{year}{2020}{\natexlab{b}}).
\newblock \bibinfo{title}{Transformation-consistent self-ensembling model for
  semi-supervised medical image segmentation}.
\newblock {\it \bibinfo{journal}{IEEE Transactions on Neural Networks and
  Learning Systems}\/},  (pp. \bibinfo{pages}{1--12}).
\bibitem[{Li et~al.(2020{\natexlab{c}})Li, Chen, Xie, Ma \& Zheng}]{li2020self}
\bibinfo{author}{Li, Y.}, \bibinfo{author}{Chen, J.}, \bibinfo{author}{Xie,
  X.}, \bibinfo{author}{Ma, K.}, \& \bibinfo{author}{Zheng, Y.}
  (\bibinfo{year}{2020}{\natexlab{c}}).
\newblock \bibinfo{title}{Self-loop uncertainty: A novel pseudo-label for
  semi-supervised medical image segmentation}.
\newblock In {\it \bibinfo{booktitle}{International Conference on Medical Image
  Computing and Computer Assisted Intervention}\/} (pp.
  \bibinfo{pages}{614--623}).
\newblock \bibinfo{organization}{Springer}.
\bibitem[{Li et~al.(2022)Li, Dan, Li, Chen, Peng, Liu \& Cai}]{li2022npcnet}
\bibinfo{author}{Li, Y.}, \bibinfo{author}{Dan, T.}, \bibinfo{author}{Li, H.},
  \bibinfo{author}{Chen, J.}, \bibinfo{author}{Peng, H.}, \bibinfo{author}{Liu,
  L.}, \& \bibinfo{author}{Cai, H.} (\bibinfo{year}{2022}).
\newblock \bibinfo{title}{{NPCNet: jointly segment primary nasopharyngeal
  carcinoma tumors and metastatic lymph nodes in MR images}}.
\newblock {\it \bibinfo{journal}{IEEE Transactions on Medical Imaging}\/},
  {\it \bibinfo{volume}{41}\/}, \bibinfo{pages}{1639--1650}.
\bibitem[{Liu et~al.(2019)Liu, Zhang, Song, Zhang, Zhang, O'Donnell \&
  Cai}]{liu2019nuclei}
\bibinfo{author}{Liu, D.}, \bibinfo{author}{Zhang, D.}, \bibinfo{author}{Song,
  Y.}, \bibinfo{author}{Zhang, C.}, \bibinfo{author}{Zhang, F.},
  \bibinfo{author}{O'Donnell, L.}, \& \bibinfo{author}{Cai, W.}
  (\bibinfo{year}{2019}).
\newblock \bibinfo{title}{{Nuclei Segmentation via a Deep Panoptic Model with
  Semantic Feature Fusion.}}
\newblock In {\it \bibinfo{booktitle}{Proceedings of the Twenty-Eighth
  International Joint Conference on Artificial Intelligence, {IJCAI-19}}\/}
  (pp. \bibinfo{pages}{861--868}).
\newblock \DOIprefix\doi{10.24963/ijcai.2019/121}.
\bibitem[{Lu et~al.(2021)Lu, Koyuncu, Corredor, Prasanna, Leo, Wang, Janowczyk,
  Bera, Lewis~Jr, Velcheti et~al.}]{lu2021feature}
\bibinfo{author}{Lu, C.}, \bibinfo{author}{Koyuncu, C.},
  \bibinfo{author}{Corredor, G.}, \bibinfo{author}{Prasanna, P.},
  \bibinfo{author}{Leo, P.}, \bibinfo{author}{Wang, X.},
  \bibinfo{author}{Janowczyk, A.}, \bibinfo{author}{Bera, K.},
  \bibinfo{author}{Lewis~Jr, J.}, \bibinfo{author}{Velcheti, V.} et~al.
  (\bibinfo{year}{2021}).
\newblock \bibinfo{title}{{Feature-driven local cell graph (FLocK): new
  computational pathology-based descriptors for prognosis of lung cancer and
  HPV status of oropharyngeal cancers}}.
\newblock {\it \bibinfo{journal}{Medical Image Analysis}\/},  {\it
  \bibinfo{volume}{68}\/}, \bibinfo{pages}{101903}.
\bibitem[{Luo et~al.(2022)Luo, Wang, Liao, Chen, Song, Chen, Zhang, Metaxas \&
  Zhang}]{luo2022semi}
\bibinfo{author}{Luo, X.}, \bibinfo{author}{Wang, G.}, \bibinfo{author}{Liao,
  W.}, \bibinfo{author}{Chen, J.}, \bibinfo{author}{Song, T.},
  \bibinfo{author}{Chen, Y.}, \bibinfo{author}{Zhang, S.},
  \bibinfo{author}{Metaxas, D.~N.}, \& \bibinfo{author}{Zhang, S.}
  (\bibinfo{year}{2022}).
\newblock \bibinfo{title}{Semi-supervised medical image segmentation via
  uncertainty rectified pyramid consistency}.
\newblock {\it \bibinfo{journal}{Medical Image Analysis}\/},  {\it
  \bibinfo{volume}{80}\/}, \bibinfo{pages}{102517}.
\bibitem[{Mehta \& Sivaswamy(2017)}]{mehta2017m}
\bibinfo{author}{Mehta, R.}, \& \bibinfo{author}{Sivaswamy, J.}
  (\bibinfo{year}{2017}).
\newblock \bibinfo{title}{{M-Net: A convolutional neural network for deep brain
  structure segmentation}}.
\newblock In {\it \bibinfo{booktitle}{2017 IEEE 14th International Symposium on
  Biomedical Imaging (ISBI 2017)}\/} (pp. \bibinfo{pages}{437--440}).
\newblock \bibinfo{organization}{IEEE}.
\bibitem[{Nielsen \& Jensen(2009)}]{nielsen2009bayesian}
\bibinfo{author}{Nielsen, T.~D.}, \& \bibinfo{author}{Jensen, F.~V.}
  (\bibinfo{year}{2009}).
\newblock {\it \bibinfo{title}{Bayesian Networks and Decision Graphs}\/}.
\newblock \bibinfo{publisher}{Springer Science \& Business Media}.
\bibitem[{Qin et~al.(2020)Qin, Zhang, Huang, Dehghan, Zaiane \&
  Jagersand}]{qin2020u2}
\bibinfo{author}{Qin, X.}, \bibinfo{author}{Zhang, Z.}, \bibinfo{author}{Huang,
  C.}, \bibinfo{author}{Dehghan, M.}, \bibinfo{author}{Zaiane, O.~R.}, \&
  \bibinfo{author}{Jagersand, M.} (\bibinfo{year}{2020}).
\newblock \bibinfo{title}{{U2-Net: Going deeper with nested U-structure for
  salient object detection}}.
\newblock {\it \bibinfo{journal}{Pattern Recognition}\/},  {\it
  \bibinfo{volume}{106}\/}, \bibinfo{pages}{107404}.
\bibitem[{Qu et~al.(2019)Qu, Yan, Riedlinger, De \& Metaxas}]{qu2019improving}
\bibinfo{author}{Qu, H.}, \bibinfo{author}{Yan, Z.},
  \bibinfo{author}{Riedlinger, G.~M.}, \bibinfo{author}{De, S.}, \&
  \bibinfo{author}{Metaxas, D.~N.} (\bibinfo{year}{2019}).
\newblock \bibinfo{title}{Improving nuclei/gland instance segmentation in
  histopathology images by full resolution neural network and spatial
  constrained loss}.
\newblock In {\it \bibinfo{booktitle}{International Conference on Medical Image
  Computing and Computer Assisted Intervention}\/} (pp.
  \bibinfo{pages}{378--386}).
\newblock \bibinfo{organization}{Springer}.
\bibitem[{Raza et~al.(2019)Raza, Cheung, Shaban, Graham, Epstein, Pelengaris,
  Khan \& Rajpoot}]{raza2019micro}
\bibinfo{author}{Raza, S. E.~A.}, \bibinfo{author}{Cheung, L.},
  \bibinfo{author}{Shaban, M.}, \bibinfo{author}{Graham, S.},
  \bibinfo{author}{Epstein, D.}, \bibinfo{author}{Pelengaris, S.},
  \bibinfo{author}{Khan, M.}, \& \bibinfo{author}{Rajpoot, N.~M.}
  (\bibinfo{year}{2019}).
\newblock \bibinfo{title}{{Micro-Net: A unified model for segmentation of
  various objects in microscopy images}}.
\newblock {\it \bibinfo{journal}{Medical Image Analysis}\/},  {\it
  \bibinfo{volume}{52}\/}, \bibinfo{pages}{160--173}.
\bibitem[{Ronneberger et~al.(2015)Ronneberger, Fischer \&
  Brox}]{ronneberger2015u}
\bibinfo{author}{Ronneberger, O.}, \bibinfo{author}{Fischer, P.}, \&
  \bibinfo{author}{Brox, T.} (\bibinfo{year}{2015}).
\newblock \bibinfo{title}{{U-Net: Convolutional networks for biomedical image
  segmentation}}.
\newblock In {\it \bibinfo{booktitle}{International Conference on Medical Image
  Computing and Computer Assisted Intervention}\/} (pp.
  \bibinfo{pages}{234--241}).
\newblock \bibinfo{organization}{Springer}.
\bibitem[{Shi et~al.(2022)Shi, Gong, Wang \& Li}]{shi2022semi}
\bibinfo{author}{Shi, J.}, \bibinfo{author}{Gong, T.}, \bibinfo{author}{Wang,
  C.}, \& \bibinfo{author}{Li, C.} (\bibinfo{year}{2022}).
\newblock \bibinfo{title}{Semi-supervised pixel contrastive learning framework
  for tissue segmentation in histopathological image}.
\newblock {\it \bibinfo{journal}{IEEE Journal of Biomedical and Health
  Informatics}\/},  {\it \bibinfo{volume}{27}\/}, \bibinfo{pages}{97--108}.
\bibitem[{Su et~al.(2015)Su, Xing, Kong, Xie, Zhang \& Yang}]{su2015robust}
\bibinfo{author}{Su, H.}, \bibinfo{author}{Xing, F.}, \bibinfo{author}{Kong,
  X.}, \bibinfo{author}{Xie, Y.}, \bibinfo{author}{Zhang, S.}, \&
  \bibinfo{author}{Yang, L.} (\bibinfo{year}{2015}).
\newblock \bibinfo{title}{Robust cell detection and segmentation in
  histopathological images using sparse reconstruction and stacked denoising
  autoencoders}.
\newblock In {\it \bibinfo{booktitle}{International Conference on Medical Image
  Computing and Computer Assisted Intervention}\/} (pp.
  \bibinfo{pages}{383--390}).
\newblock \bibinfo{organization}{Springer}.
\bibitem[{Sundaresan et~al.(2021)Sundaresan, Zamboni, Rothwell, Jenkinson \&
  Griffanti}]{sundaresan2021triplanar}
\bibinfo{author}{Sundaresan, V.}, \bibinfo{author}{Zamboni, G.},
  \bibinfo{author}{Rothwell, P.~M.}, \bibinfo{author}{Jenkinson, M.}, \&
  \bibinfo{author}{Griffanti, L.} (\bibinfo{year}{2021}).
\newblock \bibinfo{title}{{Triplanar ensemble U-Net model for white matter
  hyperintensities segmentation on MR images}}.
\newblock {\it \bibinfo{journal}{Medical Image Analysis}\/},  {\it
  \bibinfo{volume}{73}\/}, \bibinfo{pages}{102184}.
\bibitem[{Tarvainen \& Valpola(2017)}]{tarvainen2017mean}
\bibinfo{author}{Tarvainen, A.}, \& \bibinfo{author}{Valpola, H.}
  (\bibinfo{year}{2017}).
\newblock \bibinfo{title}{Mean teachers are better role models: Weight-averaged
  consistency targets improve semi-supervised deep learning results}.
\newblock In {\it \bibinfo{booktitle}{Advances in Neural Information Processing
  Systems}\/} (pp. \bibinfo{pages}{1195--1204}).
\bibitem[{Verma et~al.(2019)Verma, Lamb, Kannala, Bengio \&
  Lopez-Paz}]{verma2019interpolation}
\bibinfo{author}{Verma, V.}, \bibinfo{author}{Lamb, A.},
  \bibinfo{author}{Kannala, J.}, \bibinfo{author}{Bengio, Y.}, \&
  \bibinfo{author}{Lopez-Paz, D.} (\bibinfo{year}{2019}).
\newblock \bibinfo{title}{{Interpolation Consistency Training for
  Semi-supervised Learning}}.
\newblock In {\it \bibinfo{booktitle}{Proceedings of the 28th International
  Joint Conference on Artificial Intelligence}\/} IJCAI'19 (pp.
  \bibinfo{pages}{3635--3641}).
\newblock \bibinfo{publisher}{AAAI Press}.
\bibitem[{Wang et~al.(2019)Wang, Li, Aertsen, Deprest, Ourselin \&
  Vercauteren}]{wang2019aleatoric}
\bibinfo{author}{Wang, G.}, \bibinfo{author}{Li, W.}, \bibinfo{author}{Aertsen,
  M.}, \bibinfo{author}{Deprest, J.}, \bibinfo{author}{Ourselin, S.}, \&
  \bibinfo{author}{Vercauteren, T.} (\bibinfo{year}{2019}).
\newblock \bibinfo{title}{Aleatoric uncertainty estimation with test-time
  augmentation for medical image segmentation with convolutional neural
  networks}.
\newblock {\it \bibinfo{journal}{Neurocomputing}\/},  {\it
  \bibinfo{volume}{338}\/}, \bibinfo{pages}{34--45}.
\bibitem[{Wang et~al.(2021)Wang, Zhan, Zu, Wu, Zhou, Zhou \&
  Wang}]{wang2021tripled}
\bibinfo{author}{Wang, K.}, \bibinfo{author}{Zhan, B.}, \bibinfo{author}{Zu,
  C.}, \bibinfo{author}{Wu, X.}, \bibinfo{author}{Zhou, J.},
  \bibinfo{author}{Zhou, L.}, \& \bibinfo{author}{Wang, Y.}
  (\bibinfo{year}{2021}).
\newblock \bibinfo{title}{Tripled-uncertainty guided mean teacher model for
  semi-supervised medical image segmentation}.
\newblock In {\it \bibinfo{booktitle}{Medical Image Computing and Computer
  Assisted Intervention}\/} (pp. \bibinfo{pages}{450--460}).
\newblock \bibinfo{organization}{Springer}.
\bibitem[{Wang et~al.(2022)Wang, Zhan, Zu, Wu, Zhou, Zhou \&
  Wang}]{wang2022semi}
\bibinfo{author}{Wang, K.}, \bibinfo{author}{Zhan, B.}, \bibinfo{author}{Zu,
  C.}, \bibinfo{author}{Wu, X.}, \bibinfo{author}{Zhou, J.},
  \bibinfo{author}{Zhou, L.}, \& \bibinfo{author}{Wang, Y.}
  (\bibinfo{year}{2022}).
\newblock \bibinfo{title}{Semi-supervised medical image segmentation via a
  tripled-uncertainty guided mean teacher model with contrastive learning}.
\newblock {\it \bibinfo{journal}{Medical Image Analysis}\/},  {\it
  \bibinfo{volume}{79}\/}, \bibinfo{pages}{102447}.
\bibitem[{Wang et~al.(2023)Wang, Wang, Zhu, Fu, Li, Cheng, Feng, Li \&
  Heng}]{wang2022dual}
\bibinfo{author}{Wang, L.}, \bibinfo{author}{Wang, J.}, \bibinfo{author}{Zhu,
  L.}, \bibinfo{author}{Fu, H.}, \bibinfo{author}{Li, P.},
  \bibinfo{author}{Cheng, G.}, \bibinfo{author}{Feng, Z.}, \bibinfo{author}{Li,
  S.}, \& \bibinfo{author}{Heng, P.-A.} (\bibinfo{year}{2023}).
\newblock \bibinfo{title}{{Dual Multiscale Mean Teacher Network for
  Semi-Supervised Infection Segmentation in Chest CT Volume for COVID-19}}.
\newblock {\it \bibinfo{journal}{IEEE Transactions on Cybernetics}\/},  {\it
  \bibinfo{volume}{53}\/}, \bibinfo{pages}{6363--6375}.
\bibitem[{Wang et~al.(2020)Wang, Zhang, Tian, Zhong, Shi, Zhang \&
  He}]{wang2020double}
\bibinfo{author}{Wang, Y.}, \bibinfo{author}{Zhang, Y.}, \bibinfo{author}{Tian,
  J.}, \bibinfo{author}{Zhong, C.}, \bibinfo{author}{Shi, Z.},
  \bibinfo{author}{Zhang, Y.}, \& \bibinfo{author}{He, Z.}
  (\bibinfo{year}{2020}).
\newblock \bibinfo{title}{{Double-Uncertainty Weighted Method for
  Semi-supervised Learning}}.
\newblock In {\it \bibinfo{booktitle}{International Conference on Medical Image
  Computing and Computer Assisted Intervention}\/} (pp.
  \bibinfo{pages}{542--551}).
\newblock \bibinfo{organization}{Springer}.
\bibitem[{Wong et~al.(2018)Wong, Moradi, Tang \& Syeda-Mahmood}]{wong20183d}
\bibinfo{author}{Wong, K.~C.}, \bibinfo{author}{Moradi, M.},
  \bibinfo{author}{Tang, H.}, \& \bibinfo{author}{Syeda-Mahmood, T.}
  (\bibinfo{year}{2018}).
\newblock \bibinfo{title}{{3D segmentation with exponential logarithmic loss
  for highly unbalanced object sizes}}.
\newblock In {\it \bibinfo{booktitle}{International Conference on Medical Image
  Computing and Computer Assisted Intervention}\/} (pp.
  \bibinfo{pages}{612--619}).
\newblock \bibinfo{organization}{Springer}.
\bibitem[{Wu et~al.(2022{\natexlab{a}})Wu, Wang, Song, Yang \&
  Qin}]{wu2022cross}
\bibinfo{author}{Wu, H.}, \bibinfo{author}{Wang, Z.}, \bibinfo{author}{Song,
  Y.}, \bibinfo{author}{Yang, L.}, \& \bibinfo{author}{Qin, J.}
  (\bibinfo{year}{2022}{\natexlab{a}}).
\newblock \bibinfo{title}{Cross-patch dense contrastive learning for
  semi-supervised segmentation of cellular nuclei in histopathologic images}.
\newblock In {\it \bibinfo{booktitle}{Proceedings of the IEEE/CVF Conference on
  Computer Vision and Pattern Recognition}\/} (pp.
  \bibinfo{pages}{11666--11675}).
\bibitem[{Wu et~al.(2022{\natexlab{b}})Wu, Ge, Zhang, Xu, Zhang, Xia \&
  Cai}]{wu2022mutual}
\bibinfo{author}{Wu, Y.}, \bibinfo{author}{Ge, Z.}, \bibinfo{author}{Zhang,
  D.}, \bibinfo{author}{Xu, M.}, \bibinfo{author}{Zhang, L.},
  \bibinfo{author}{Xia, Y.}, \& \bibinfo{author}{Cai, J.}
  (\bibinfo{year}{2022}{\natexlab{b}}).
\newblock \bibinfo{title}{Mutual consistency learning for semi-supervised
  medical image segmentation}.
\newblock {\it \bibinfo{journal}{Medical Image Analysis}\/},  {\it
  \bibinfo{volume}{81}\/}, \bibinfo{pages}{102530}.
\bibitem[{Xiang et~al.(2020)Xiang, Zhang, Liu, Song, Huang \&
  Cai}]{xiang2020bio}
\bibinfo{author}{Xiang, T.}, \bibinfo{author}{Zhang, C.}, \bibinfo{author}{Liu,
  D.}, \bibinfo{author}{Song, Y.}, \bibinfo{author}{Huang, H.}, \&
  \bibinfo{author}{Cai, W.} (\bibinfo{year}{2020}).
\newblock \bibinfo{title}{{BiO-Net: Learning Recurrent Bi-directional
  Connections for Encoder-Decoder Architecture}}.
\newblock In {\it \bibinfo{booktitle}{International Conference on Medical Image
  Computing and Computer Assisted Intervention}\/} (pp.
  \bibinfo{pages}{74--84}).
\newblock \bibinfo{organization}{Springer}.
\bibitem[{Xiang et~al.(2022)Xiang, Zhang, Wang, Song, Liu, Huang \&
  Cai}]{XIANG2022102420}
\bibinfo{author}{Xiang, T.}, \bibinfo{author}{Zhang, C.},
  \bibinfo{author}{Wang, X.}, \bibinfo{author}{Song, Y.}, \bibinfo{author}{Liu,
  D.}, \bibinfo{author}{Huang, H.}, \& \bibinfo{author}{Cai, W.}
  (\bibinfo{year}{2022}).
\newblock \bibinfo{title}{Towards bi-directional skip connections in
  encoder-decoder architectures and beyond}.
\newblock {\it \bibinfo{journal}{Medical Image Analysis}\/},  {\it
  \bibinfo{volume}{78}\/}, \bibinfo{pages}{102420}.
\bibitem[{Xie et~al.(2019)Xie, Lu, Zhang, Shen \& Xia}]{xie2019deep}
\bibinfo{author}{Xie, Y.}, \bibinfo{author}{Lu, H.}, \bibinfo{author}{Zhang,
  J.}, \bibinfo{author}{Shen, C.}, \& \bibinfo{author}{Xia, Y.}
  (\bibinfo{year}{2019}).
\newblock \bibinfo{title}{{Deep Segmentation-Emendation Model for Gland
  Instance Segmentation}}.
\newblock In {\it \bibinfo{booktitle}{International Conference on Medical Image
  Computing and Computer Assisted Intervention}\/} (pp.
  \bibinfo{pages}{469--477}).
\newblock \bibinfo{organization}{Springer}.
\bibitem[{Xie et~al.(2020)Xie, Zhang, Liao, Verjans, Shen \&
  Xia}]{xie2020pairwise}
\bibinfo{author}{Xie, Y.}, \bibinfo{author}{Zhang, J.}, \bibinfo{author}{Liao,
  Z.}, \bibinfo{author}{Verjans, J.}, \bibinfo{author}{Shen, C.}, \&
  \bibinfo{author}{Xia, Y.} (\bibinfo{year}{2020}).
\newblock \bibinfo{title}{{Pairwise Relation Learning for Semi-supervised Gland
  Segmentation}}.
\newblock In {\it \bibinfo{booktitle}{International Conference on Medical Image
  Computing and Computer Assisted Intervention}\/} (pp.
  \bibinfo{pages}{417--427}).
\newblock \bibinfo{organization}{Springer}.
\bibitem[{Xu et~al.(2022)Xu, Wang, Zhang, Han, Zhang, Chen \&
  Li}]{xu2022bmanet}
\bibinfo{author}{Xu, C.}, \bibinfo{author}{Wang, Y.}, \bibinfo{author}{Zhang,
  D.}, \bibinfo{author}{Han, L.}, \bibinfo{author}{Zhang, Y.},
  \bibinfo{author}{Chen, J.}, \& \bibinfo{author}{Li, S.}
  (\bibinfo{year}{2022}).
\newblock \bibinfo{title}{{BMAnet: Boundary mining with adversarial learning
  for semi-supervised 2D myocardial infarction segmentation}}.
\newblock {\it \bibinfo{journal}{IEEE Journal of Biomedical and Health
  Informatics}\/},  {\it \bibinfo{volume}{27}\/}, \bibinfo{pages}{87--96}.
\bibitem[{Xu et~al.(2021)Xu, Lian, Wang, Zhu, Chen, Wang, Royce, Yap, Shen \&
  Lian}]{xu2021asymmetric}
\bibinfo{author}{Xu, X.}, \bibinfo{author}{Lian, C.}, \bibinfo{author}{Wang,
  S.}, \bibinfo{author}{Zhu, T.}, \bibinfo{author}{Chen, R.~C.},
  \bibinfo{author}{Wang, A.~Z.}, \bibinfo{author}{Royce, T.~J.},
  \bibinfo{author}{Yap, P.-T.}, \bibinfo{author}{Shen, D.}, \&
  \bibinfo{author}{Lian, J.} (\bibinfo{year}{2021}).
\newblock \bibinfo{title}{Asymmetric multi-task attention network for prostate
  bed segmentation in computed tomography images}.
\newblock {\it \bibinfo{journal}{Medical Image Analysis}\/},  {\it
  \bibinfo{volume}{72}\/}, \bibinfo{pages}{102116}.
\bibitem[{Xu et~al.(2023)Xu, Wang, Lu, Luo, Yan, Zheng \&
  Tong}]{xu2023ambiguity}
\bibinfo{author}{Xu, Z.}, \bibinfo{author}{Wang, Y.}, \bibinfo{author}{Lu, D.},
  \bibinfo{author}{Luo, X.}, \bibinfo{author}{Yan, J.}, \bibinfo{author}{Zheng,
  Y.}, \& \bibinfo{author}{Tong, R. K.-y.} (\bibinfo{year}{2023}).
\newblock \bibinfo{title}{Ambiguity-selective consistency regularization for
  mean-teacher semi-supervised medical image segmentation}.
\newblock {\it \bibinfo{journal}{Medical Image Analysis}\/},  {\it
  \bibinfo{volume}{88}\/}, \bibinfo{pages}{102880}.
\bibitem[{Yan et~al.(2023{\natexlab{a}})Yan, Guo \& Liu}]{yan2023pretp}
\bibinfo{author}{Yan, K.}, \bibinfo{author}{Guo, Y.}, \& \bibinfo{author}{Liu,
  B.} (\bibinfo{year}{2023}{\natexlab{a}}).
\newblock \bibinfo{title}{{PreTP-2L: Identification of therapeutic peptides and
  their types using two-layer ensemble learning framework}}.
\newblock {\it \bibinfo{journal}{Bioinformatics}\/},  {\it
  \bibinfo{volume}{39}\/}, \bibinfo{pages}{btad125}.
\bibitem[{Yan et~al.(2023{\natexlab{b}})Yan, Lv, Guo, Peng \&
  Liu}]{yan2023samppred}
\bibinfo{author}{Yan, K.}, \bibinfo{author}{Lv, H.}, \bibinfo{author}{Guo, Y.},
  \bibinfo{author}{Peng, W.}, \& \bibinfo{author}{Liu, B.}
  (\bibinfo{year}{2023}{\natexlab{b}}).
\newblock \bibinfo{title}{{sAMPpred-GAT: prediction of antimicrobial peptide by
  graph attention network and predicted peptide structure}}.
\newblock {\it \bibinfo{journal}{Bioinformatics}\/},  {\it
  \bibinfo{volume}{39}\/}, \bibinfo{pages}{btac715}.
\bibitem[{Yang et~al.(2023)Yang, Dasmahapatra \& Mahmoodi}]{yang2023ads_unet}
\bibinfo{author}{Yang, Y.}, \bibinfo{author}{Dasmahapatra, S.}, \&
  \bibinfo{author}{Mahmoodi, S.} (\bibinfo{year}{2023}).
\newblock \bibinfo{title}{{ADS\_UNet: A nested UNet for histopathology image
  segmentation}}.
\newblock {\it \bibinfo{journal}{Expert Systems with Applications}\/},  {\it
  \bibinfo{volume}{226}\/}, \bibinfo{pages}{120128}.
\bibitem[{Yang \& Farsiu(2023)}]{yang2023directional}
\bibinfo{author}{Yang, Z.}, \& \bibinfo{author}{Farsiu, S.}
  (\bibinfo{year}{2023}).
\newblock \bibinfo{title}{Directional connectivity-based segmentation of
  medical images}.
\newblock In {\it \bibinfo{booktitle}{Proceedings of the IEEE/CVF Conference on
  Computer Vision and Pattern Recognition}\/} (pp.
  \bibinfo{pages}{11525--11535}).
\bibitem[{Yu et~al.(2020)Yu, Wang, Gao, Yu, Shen \& Sang}]{yu2020context}
\bibinfo{author}{Yu, C.}, \bibinfo{author}{Wang, J.}, \bibinfo{author}{Gao,
  C.}, \bibinfo{author}{Yu, G.}, \bibinfo{author}{Shen, C.}, \&
  \bibinfo{author}{Sang, N.} (\bibinfo{year}{2020}).
\newblock \bibinfo{title}{Context prior for scene segmentation}.
\newblock In {\it \bibinfo{booktitle}{Proceedings of the IEEE/CVF Conference on
  Computer Vision and Pattern Recognition}\/} (pp.
  \bibinfo{pages}{12416--12425}).
\bibitem[{Yu \& Koltun(2015)}]{yu2015multi}
\bibinfo{author}{Yu, F.}, \& \bibinfo{author}{Koltun, V.}
  (\bibinfo{year}{2015}).
\newblock \bibinfo{title}{Multi-scale context aggregation by dilated
  convolutions}.
\newblock {\it \bibinfo{journal}{arXiv preprint arXiv:1511.07122}\/}, .
\bibitem[{Yu et~al.(2019)Yu, Wang, Li, Fu \& Heng}]{yu2019uncertainty}
\bibinfo{author}{Yu, L.}, \bibinfo{author}{Wang, S.}, \bibinfo{author}{Li, X.},
  \bibinfo{author}{Fu, C.-W.}, \& \bibinfo{author}{Heng, P.-A.}
  (\bibinfo{year}{2019}).
\newblock \bibinfo{title}{{Uncertainty-aware self-ensembling model for
  semi-supervised 3D left atrium segmentation}}.
\newblock In {\it \bibinfo{booktitle}{International Conference on Medical Image
  Computing and Computer Assisted Intervention}\/} (pp.
  \bibinfo{pages}{605--613}).
\newblock \bibinfo{organization}{Springer}.
\bibitem[{Yu et~al.(2023)Yu, Yang, Zhou, Cai, Gao, Lee, Li, Bao, Xu, Lasko
  et~al.}]{yu2023unest}
\bibinfo{author}{Yu, X.}, \bibinfo{author}{Yang, Q.}, \bibinfo{author}{Zhou,
  Y.}, \bibinfo{author}{Cai, L.~Y.}, \bibinfo{author}{Gao, R.},
  \bibinfo{author}{Lee, H.~H.}, \bibinfo{author}{Li, T.}, \bibinfo{author}{Bao,
  S.}, \bibinfo{author}{Xu, Z.}, \bibinfo{author}{Lasko, T.~A.} et~al.
  (\bibinfo{year}{2023}).
\newblock \bibinfo{title}{{UNesT: local spatial representation learning with
  hierarchical transformer for efficient medical segmentation}}.
\newblock {\it \bibinfo{journal}{Medical Image Analysis}\/},  {\it
  \bibinfo{volume}{90}\/}, \bibinfo{pages}{102939}.
\bibitem[{Zhang et~al.(2019)Zhang, Xie, Xia \& Shen}]{zhang2019attention}
\bibinfo{author}{Zhang, J.}, \bibinfo{author}{Xie, Y.}, \bibinfo{author}{Xia,
  Y.}, \& \bibinfo{author}{Shen, C.} (\bibinfo{year}{2019}).
\newblock \bibinfo{title}{{Attention Residual Learning for Skin Lesion
  Classification}}.
\newblock {\it \bibinfo{journal}{IEEE Transactions on Medical Imaging}\/},
  {\it \bibinfo{volume}{38}\/}, \bibinfo{pages}{2092--2103}.
\bibitem[{Zhang et~al.(2023)Zhang, Zhang, Tian, Lukasiewicz \&
  Xu}]{zhang2023multi}
\bibinfo{author}{Zhang, S.}, \bibinfo{author}{Zhang, J.},
  \bibinfo{author}{Tian, B.}, \bibinfo{author}{Lukasiewicz, T.}, \&
  \bibinfo{author}{Xu, Z.} (\bibinfo{year}{2023}).
\newblock \bibinfo{title}{Multi-modal contrastive mutual learning and
  pseudo-label re-learning for semi-supervised medical image segmentation}.
\newblock {\it \bibinfo{journal}{Medical Image Analysis}\/},  {\it
  \bibinfo{volume}{83}\/}, \bibinfo{pages}{102656}.
\bibitem[{Zhang et~al.(2017)Zhang, Yang, Chen, Fredericksen, Hughes \&
  Chen}]{zhang2017deep}
\bibinfo{author}{Zhang, Y.}, \bibinfo{author}{Yang, L.}, \bibinfo{author}{Chen,
  J.}, \bibinfo{author}{Fredericksen, M.}, \bibinfo{author}{Hughes, D.~P.}, \&
  \bibinfo{author}{Chen, D.~Z.} (\bibinfo{year}{2017}).
\newblock \bibinfo{title}{Deep adversarial networks for biomedical image
  segmentation utilizing unannotated images}.
\newblock In {\it \bibinfo{booktitle}{International Conference on Medical Image
  Computing and Computer Assisted Intervention}\/} (pp.
  \bibinfo{pages}{408--416}).
\newblock \bibinfo{organization}{Springer}.
\bibitem[{Zhao et~al.(2020)Zhao, Chen, Li, Yu, Yao, Yan, Wang, Liu, Liang \&
  Han}]{ZHAO2020101786}
\bibinfo{author}{Zhao, B.}, \bibinfo{author}{Chen, X.}, \bibinfo{author}{Li,
  Z.}, \bibinfo{author}{Yu, Z.}, \bibinfo{author}{Yao, S.},
  \bibinfo{author}{Yan, L.}, \bibinfo{author}{Wang, Y.}, \bibinfo{author}{Liu,
  Z.}, \bibinfo{author}{Liang, C.}, \& \bibinfo{author}{Han, C.}
  (\bibinfo{year}{2020}).
\newblock \bibinfo{title}{{Triple U-Net: Hematoxylin-aware nuclei segmentation
  with progressive dense feature aggregation}}.
\newblock {\it \bibinfo{journal}{Medical Image Analysis}\/},  {\it
  \bibinfo{volume}{65}\/}, \bibinfo{pages}{101786}.
\bibitem[{Zhao et~al.(2023)Zhao, Qi, Wang, Wang, Wu, Mao \&
  Zhang}]{zhao2023rcps}
\bibinfo{author}{Zhao, X.}, \bibinfo{author}{Qi, Z.}, \bibinfo{author}{Wang,
  S.}, \bibinfo{author}{Wang, Q.}, \bibinfo{author}{Wu, X.},
  \bibinfo{author}{Mao, Y.}, \& \bibinfo{author}{Zhang, L.}
  (\bibinfo{year}{2023}).
\newblock \bibinfo{title}{{RCPS: Rectified contrastive pseudo supervision for
  semi-supervised medical image segmentation}}.
\newblock {\it \bibinfo{journal}{arXiv preprint arXiv:2301.05500}\/}, .
\bibitem[{Zheng et~al.(2020)Zheng, Motch~Perrine, Pitirri, Kawasaki, Wang,
  Richtsmeier \& Chen}]{zheng2020cartilage}
\bibinfo{author}{Zheng, H.}, \bibinfo{author}{Motch~Perrine, S.~M.},
  \bibinfo{author}{Pitirri, M.~K.}, \bibinfo{author}{Kawasaki, K.},
  \bibinfo{author}{Wang, C.}, \bibinfo{author}{Richtsmeier, J.~T.}, \&
  \bibinfo{author}{Chen, D.~Z.} (\bibinfo{year}{2020}).
\newblock \bibinfo{title}{{Cartilage segmentation in high-resolution 3D
  micro-CT images via uncertainty-guided self-training with very sparse
  annotation}}.
\newblock In {\it \bibinfo{booktitle}{International Conference on Medical Image
  Computing and Computer Assisted Intervention}\/} (pp.
  \bibinfo{pages}{802--812}).
\newblock \bibinfo{organization}{Springer}.
\bibitem[{Zheng et~al.(2019)Zheng, Zhang, Yang, Liang, Zhao, Wang \&
  Chen}]{zheng2019new}
\bibinfo{author}{Zheng, H.}, \bibinfo{author}{Zhang, Y.},
  \bibinfo{author}{Yang, L.}, \bibinfo{author}{Liang, P.},
  \bibinfo{author}{Zhao, Z.}, \bibinfo{author}{Wang, C.}, \&
  \bibinfo{author}{Chen, D.~Z.} (\bibinfo{year}{2019}).
\newblock \bibinfo{title}{{A new ensemble learning framework for 3D biomedical
  image segmentation}}.
\newblock In {\it \bibinfo{booktitle}{Proceedings of the AAAI Conference on
  Artificial Intelligence}\/} (pp. \bibinfo{pages}{5909--5916}).
\newblock volume~\bibinfo{volume}{33}.
\bibitem[{Zheng et~al.(2022)Zheng, Xu \& Wei}]{zheng2022double}
\bibinfo{author}{Zheng, K.}, \bibinfo{author}{Xu, J.}, \& \bibinfo{author}{Wei,
  J.} (\bibinfo{year}{2022}).
\newblock \bibinfo{title}{Double noise mean teacher self-ensembling model for
  semi-supervised tumor segmentation}.
\newblock In {\it \bibinfo{booktitle}{ICASSP 2022-2022 IEEE International
  Conference on Acoustics, Speech and Signal Processing (ICASSP)}\/} (pp.
  \bibinfo{pages}{1446--1450}).
\newblock \bibinfo{organization}{IEEE}.
\bibitem[{Zheng \& Yang(2021)}]{zheng2021rectifying}
\bibinfo{author}{Zheng, Z.}, \& \bibinfo{author}{Yang, Y.}
  (\bibinfo{year}{2021}).
\newblock \bibinfo{title}{Rectifying pseudo label learning via uncertainty
  estimation for domain adaptive semantic segmentation}.
\newblock {\it \bibinfo{journal}{International Journal of Computer Vision}\/},
  {\it \bibinfo{volume}{129}\/}, \bibinfo{pages}{1106--1120}.
\bibitem[{Zhong et~al.(2020)Zhong, Lin, Bidart, Hu, Daya, Li, Zheng, Li \&
  Wong}]{zhong2020squeeze}
\bibinfo{author}{Zhong, Z.}, \bibinfo{author}{Lin, Z.~Q.},
  \bibinfo{author}{Bidart, R.}, \bibinfo{author}{Hu, X.},
  \bibinfo{author}{Daya, I.~B.}, \bibinfo{author}{Li, Z.},
  \bibinfo{author}{Zheng, W.-S.}, \bibinfo{author}{Li, J.}, \&
  \bibinfo{author}{Wong, A.} (\bibinfo{year}{2020}).
\newblock \bibinfo{title}{Squeeze-and-attention networks for semantic
  segmentation}.
\newblock In {\it \bibinfo{booktitle}{Proceedings of the IEEE/CVF Conference on
  Computer Vision and Pattern Recognition}\/} (pp.
  \bibinfo{pages}{13065--13074}).
\bibitem[{Zhou et~al.(2020)Zhou, Chen, Lin \& Heng}]{zhou2020deep}
\bibinfo{author}{Zhou, Y.}, \bibinfo{author}{Chen, H.}, \bibinfo{author}{Lin,
  H.}, \& \bibinfo{author}{Heng, P.-A.} (\bibinfo{year}{2020}).
\newblock \bibinfo{title}{{Deep Semi-supervised Knowledge Distillation for
  Overlapping Cervical Cell Instance Segmentation}}.
\newblock In {\it \bibinfo{booktitle}{International Conference on Medical Image
  Computing and Computer Assisted Intervention}\/} (pp.
  \bibinfo{pages}{521--531}).
\newblock \bibinfo{organization}{Springer}.
\bibitem[{Zhu et~al.(2023)Zhu, Bolsterlee, Chow, Song \&
  Meijering}]{zhu2023hybrid}
\bibinfo{author}{Zhu, J.}, \bibinfo{author}{Bolsterlee, B.},
  \bibinfo{author}{Chow, B.~V.}, \bibinfo{author}{Song, Y.}, \&
  \bibinfo{author}{Meijering, E.} (\bibinfo{year}{2023}).
\newblock \bibinfo{title}{{Hybrid Dual Mean-Teacher Network With
  Double-Uncertainty Guidance for Semi-Supervised Segmentation of MRI Scans}}.
\newblock {\it \bibinfo{journal}{arXiv preprint arXiv:2303.05126}\/}, .

\end{thebibliography}
\end{document}